\documentclass[acmlarge]{acmart}
\usepackage{graphicx}
\usepackage{multirow}
\usepackage{hyperref}
\usepackage{amsthm}
\usepackage{booktabs}
\usepackage{subcaption}
\usepackage{makecell}
\usepackage[inkscapelatex=false]{svg}
\usepackage{microtype}

\usepackage{color}
\usepackage{multirow}
\usepackage{makecell}
\usepackage{comment}
\usepackage{threeparttable}
\usepackage[edges]{forest}

\theoremstyle{definition}
\newtheorem{definition}{Definition}[section]

\theoremstyle{definition}
\newtheorem{desc}{Description}[section]

\newcommand{\red}[1]{{\textcolor{red}{#1}}}
\newcommand{\blue}[1]{{\textcolor{blue}{#1}}}
\newcommand{\orange}[1]{{\textcolor{orange}{#1}}}

\begin{document}

\title{Natural Language Reasoning, A Survey}

\author{Fei Yu, Hongbo Zhang}
\affiliation{%
  \institution{The Chinese University of Hong Kong, Shenzhen}
  \city{Shenzhen}
  \country{China}
}
\email{feiyu1@link.cuhk.edu.cn,hongboz183@gmail.com}

\author{Prayag Tiwari}
\affiliation{%
  \institution{School of Information Technology, Halmstad University}
  \city{Halmstad}
  \country{Sweden}
}
\email{prayag.tiwari@ieee.org}

\author{Benyou Wang}
\authornote{Corresponding author}
\affiliation{%
  \institution{The Chinese University of Hong Kong, Shenzhen}
  \city{Shenzhen}
  \country{China}
}
\email{wangbenyou@cuhk.edu.cn}

\begin{abstract}
This survey paper proposes a clearer view of natural language reasoning in the field of Natural Language Processing (NLP), both conceptually and practically. Conceptually, we provide a distinct definition for natural language reasoning in NLP, based on both philosophy and NLP scenarios, discuss what types of tasks require reasoning, and introduce a taxonomy of reasoning. Practically, we conduct a comprehensive literature review on natural language reasoning in NLP, mainly covering classical logical reasoning, natural language inference, multi-hop question answering, and commonsense reasoning. The paper also identifies and views backward reasoning, a powerful paradigm for multi-step reasoning, and introduces defeasible reasoning as one of the most important future directions in natural language reasoning research. We focus on single-modality unstructured natural language text, excluding neuro-symbolic techniques and mathematical reasoning\footnote{https://github.com/FreedomIntelligence/ReasoningNLP}.
\end{abstract}

\maketitle

\ccsdesc[500]{Computing methodologies~Natural language processing}
\ccsdesc[300]{Computing methodologies~Machine learning approaches}

\keywords{Natural language reasoning, pre-trained language models}

\newpage

\setcounter{secnumdepth}{3}
\setcounter{tocdepth}{3}

\newpage

\section{Introduction}

Natural Language Processing (NLP) has shown significant advancements in recent years, particularly with the introduction of transformers and pre-trained language models (PLMs). However, their abilities\footnote{In this survey, we refer to transformer-based pre-trained language models.} to perform natural language reasoning (NLR) are still far from satisfactory. Reasoning, the process of making inferences based on existing knowledge, is a fundamental aspect of human intelligence and is essential for complex tasks such as decision-making. Building an artificial intelligence system capable of reasoning is both the ultimate goal of the research community and the necessary way to improve the performance of complex applications. Compared to reason with formal language, reasoning with natural language expressions provides a more natural human-computer interaction interface and opens the door to research on defeasible reasoning, such as abduction and induction, which are incapable of formal-based symbolic methods. 

PLMs such as BERT~\cite{BERT19} and GPT~\cite{GPT-18} have been the essential components in NLP research since they occurred. Pre-trained on large-scale text corpora, PLMs are capable of natural language understanding. Recent progresses suggest that PLMs also have the potential to solve reasoning problems~\cite{ruletaker20, LeapOfThoughtT20, COT22, BBH22}. Specifically, PLMs can perform soft deductive reasoning over natural language statements~\cite{ruletaker20}, reason with implicit knowledge memorized in their parameters~\cite{LeapOfThoughtT20}, and perform multi-step reasoning step-by-step just with a few demonstrations or instructions when the model size is large enough via chain-of-thought prompting~\cite{COT22, LLM-ZSR22}. Recently, ChatGPT and GPT4 also made impressive reasoning capabilities to the community~\cite{chatgpt_eva23,sparks23}.

However, while reasoning has attracted increasing attention recently~\cite{ruletaker20, COT22, LLM-ZSR22,selection-inference22,faithful22, Entailer22,self-ask22}, there still lacks a distinct definition of reasoning and the term ``reasoning'' is sometimes of mistaken usage, which may affect the communication and development towards reasoning in the NLP community. For example, while it belongs to ``commonsense reasoning'', few people might deem that telling about a shared lived experiences~\cite{ProtoQA20}, e.g. ``name something that you might forget in a hotel room'', is reasoning. Another example is that sometimes ``natural language inference'' is introduced as a task of natural language understanding~\cite{SNLI15}, but other times of reasoning~\cite{ruletaker20}. By now, none all of the tasks named with ``reasoning'' are believed as reasoning (e.g. commonsense reasoning), and none all of the tasks named ``without reasoning'' are thought of as non-reasoning (e.g. natural language inference and multi-hop question answering). This raises a question: what reasoning is actually, and how can we identify reasoning tasks if their names are not much indicative? Although many researches~\cite{surveyHuang22,ruletaker20,AbductionRules22,DEER22} refer to a definition of reasoning from philosophy and logic, the definition cannot capture the reasoning in NLP well enough. For example, while reasoning is philosophically defined as ``using evidence and logic to arrive at conclusions''~\cite{surveyHuang22}, it fails to clarify whether implicit commonsense knowledge can be evidence and what types of conclusion are reasoning products, e.g. how about named-entity disambiguation?

To promote the research on reasoning in NLP, we make an attempt to propose a clearer view of NLP reasoning, both conceptually and practically. Conceptually, we propose a definition for NLP reasoning based on both philosophy and NLP scenarios, discuss what types of tasks require reasoning, and introduce a taxonomy of reasoning. Practically, we provide a comprehensive literature review on natural language reasoning in NLP based on our clarified definition, mainly covering classical logical reasoning, natural language inference, multi-hop question answering, and commonsense reasoning. Reviewing papers of all sizes of PLMs, we capture general methodologies that can be applied to different model sizes: end-to-end reasoning, forward reasoning, and backward reasoning. Finally, we discuss some limitations and future directions of reasoning.

In addition to the definition of reasoning, there is an important point distinguishing this survey from the other surveys~\cite{surveyHuang22,surveyQiao22,surveyYang22}: we identify and view backward reasoning, another powerful paradigm for multi-step reasoning in addition to forward reasoning. While forward reasoning, such as chain-of-thought prompting, has been popular in LLMs recently, we argue that it is worth conducting more exploration of backward reasoning. Backward reasoning is more efficient than forward reasoning both conceptually and empirically due to smaller search space~\cite{LAMBADA22}, which is the potential to generalize to complex reasoning with longer steps. 

\begin{figure}[ht]
\centering\small
\resizebox{0.90\textwidth}{!}{




    
\definecolor{green}{HTML}{70AD47}
\definecolor{orange}{HTML}{ED7D31}
\definecolor{blue}{HTML}{5B9BD5}
\definecolor{yellow}{HTML}{FFC000}

\tikzset{
    basic/.style  = {draw, text width=2cm, align=center, font=\sffamily, rectangle},
    root/.style   = {basic, thin, rounded corners=2pt, thin, align=center, fill=orange!80},
    onode/.style = {basic, thin, rounded corners=2pt, align=center, fill=yellow!80,text width=2cm,},
    tnode/.style = {basic, thin, rounded corners=2pt, align=left, fill=green!80, text width=4cm, align=center},
    xnode/.style = {basic, thin, rounded corners=2pt, align=center, fill=blue!80,text width=3cm,},
    wnode/.style = {basic, thin, rounded corners=2pt, align=left, fill=blue!80, text width=2cm},
    edge from parent/.style={draw=black, edge from parent fork right}

}

\begin{forest} 
for tree={
    grow=east,
    growth parent anchor=west,
    reversed = true,
    parent anchor=east,
    child anchor=west,
    edge path={\noexpand\path[\forestoption{edge},->, >={latex}] 
         (!u.parent anchor) -- +(10pt,0pt) |-  (.child anchor) 
         \forestoption{edge label};}
}
[
    Reasoning and PLMs, root,  l sep=10mm,
    [\S~\ref{sec:what}: What is Reasoning, onode,  l sep=10mm,
        [Definition, tnode]
        [Categories, tnode]
        [Potential\& Challenges \& Requirements, tnode]
    ]
    [\S~\ref{sec:why}: Why PLMs for Reasoning, onode,  l sep=10mm,
        [Introduction to PLMs, tnode
        ]
        [Empirical Development, tnode]
    ]
    [\S~\ref{sec:method}: Methodologies of NLR, onode,  l sep=10mm,
        [End-to-End Reasoning, tnode
        ]
        [Forward Reasoning, tnode
        ]
        [Basckward Reasoning, tnode
        ]
        [Summary , tnode]
    ]
    [\S~\ref{sec:topics}: NLR Benchmarks, onode,  l sep=10mm,
        [Classical Logical Reasoning, tnode
        ]
        [Natural Language Inference, tnode, l sep=10mm]
        [Multi-Hop QA, tnode, l sep=10mm]
        [Commonsense Reasoning, tnode, l sep=10mm]
        [Complex Reasoning, tnode, l sep=10mm]
        [Others, tnode, l sep=10mm]
    ]
    [\S~\ref{sec:discuss}: Discussion, onode,  l sep=10mm,
        [Open Question, tnode, l sep=10mm]
        [Limitations, tnode, l sep=10mm]
        [Future, tnode, l sep=10mm]
    ]
]
\end{forest}


}
\caption{Architecture of this survey.}
\label{fig:architecture}
\end{figure}
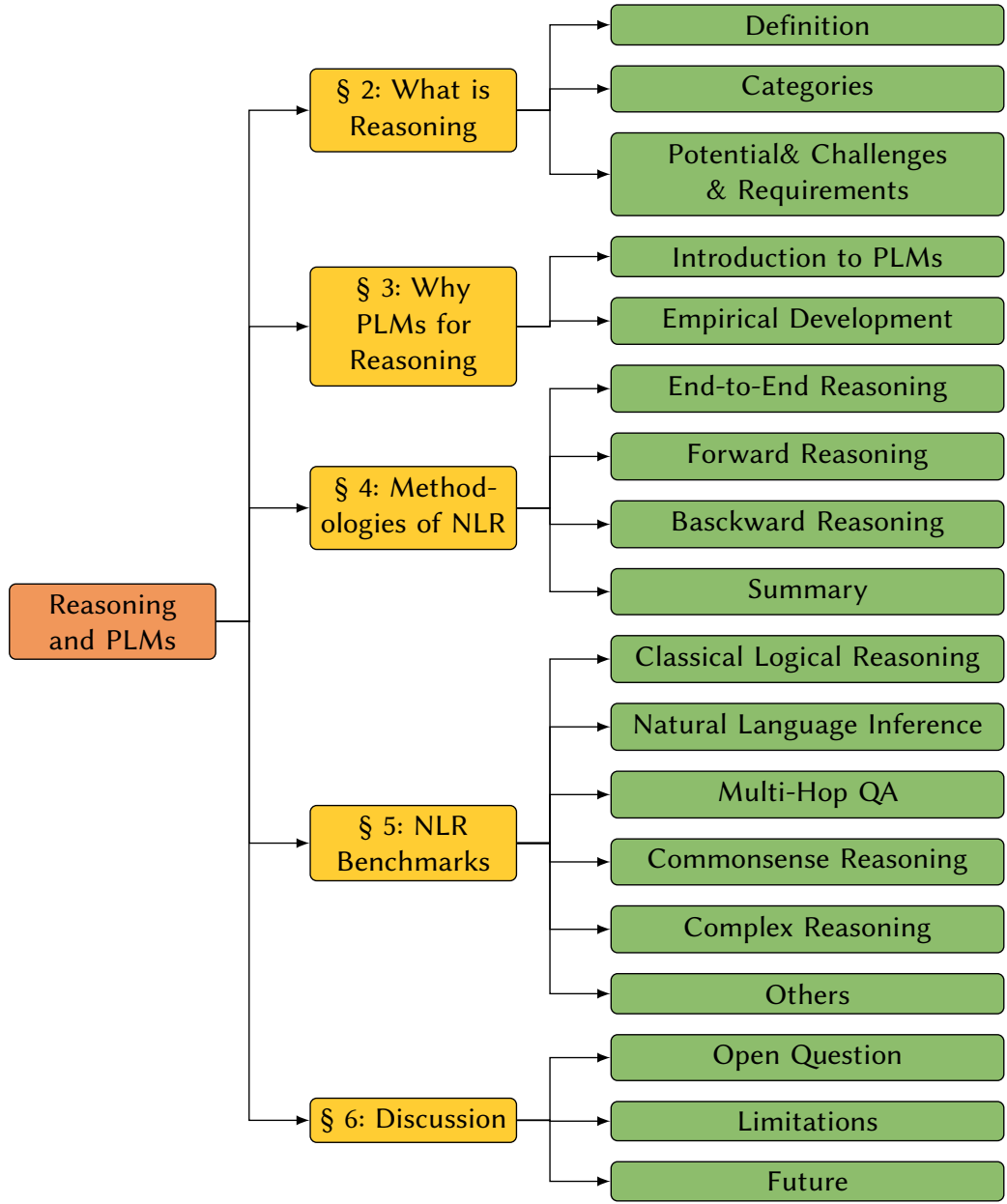

In this article, we focus on the single-modality unstructured natural language text (without knowledge triples, tables and intermediate formal language) and natural language reasoning (rather than symbolic reasoning and mathematical reasoning)\footnote{Although recently it is popular to solve mathematical reasoning problems such as math word problems using NLP methods, we do not cover them in this paper since mathematical reasoning is very different to natural language reasoning in nature as math is precise and formal.}. Concretely, We conduct a review of related works that utilize transformer-based PLMs, with a deliberate exclusion of neuro-symbolic techniques. We sorted the collected papers and categorised the methodologies of natural language reasoning in NLP. We identify the progress and trend in recent years in this domain. The paper is organized into five sections (as shown in Figure~\ref{fig:architecture}). 


We collected more than two hundred papers related to reasoning or PLMs in recent years. We searched keywords such as inference, reasoning, infer, reason, multi-step, and multi-hop on the top conferences, including ACL, EMNLP, NAACL, ICML, ICLR, and NeurIPS, from 2019 to 2022. We also found some related works from the collected papers. 


\begin{figure}[ht]
\centering\scriptsize
\includegraphics[width=0.92\linewidth]{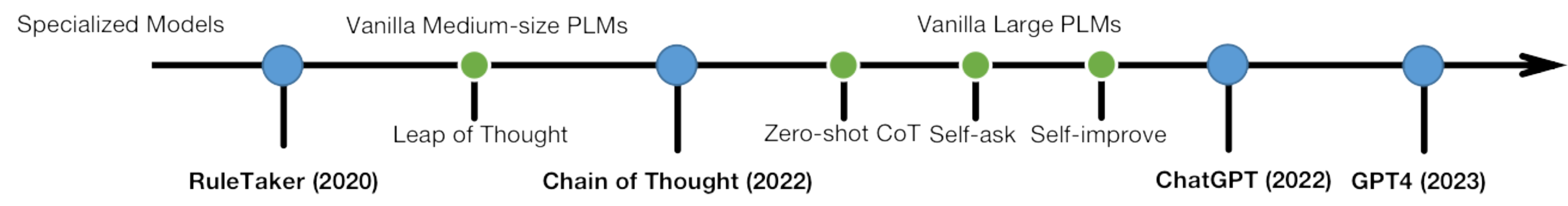}
\caption{Timeline of important works.}
\label{fig:timeline of Popular Methods}
\end{figure}

In conclusion, the main contributions of this survey are 
\begin{enumerate}
    \item To our best knowledge, we are the first to provide a distinct definition for natural language reasoning in NLP and discuss to what degree some popular benchmarks are related to reasoning.
    \item To our best knowledge, we are the first to conduct a comprehensive review on PLM-based natural language reasoning, covering diverse NLR benchmarks, and providing a comprehensive taxonomy of methodology. We also cover backward reasoning, which is neglected but has potential. 
    \item We introduce defeasible reasoning, which we believe is one of the most potential future directions, compare differences between deductive reasoning and defeasible reasoning, discuss how they can affect NLP solutions, and review current methods.
\end{enumerate}

\section{What is Natural Language Reasoning}\label{sec:what}
There still lacks a distinct definition of natural language reasoning in NLP, which affects the development and communication of NLR in the NLP community. To promote understanding, analysis and communication, we aim to suggest distinct definitions of terms and concepts for natural language reasoning in NLP. To realize this goal, we take a look into two relevant areas which have studied reasoning for a long time: philosophy and logic and transfer the relevant reasoning theory into NLP. First, we propose a definition for NLR in NLP that satisfies the concerns of the NLP community (Sec~\ref{sec:def}). Then, we provide categories of NLR and introduce how the differences between them can affect NLP solutions (Sec~\ref{sec:category}). Finally, we introduce the potentials, challenges, and requirements to achieve NLR (Sec~\ref{sec:nlr_pc}).

\subsection{Definition}\label{sec:def}
Reasoning in NLP has been focused on in recent years while philosophy has studied reasoning since thousand years ago, and logic is seen as the art of correct reasoning, which studies the concepts of inference, systematizes its categories, and develops principles of good reasoning, including formal logic and informal logic~\cite{Oxford08,hurley14,goldman86}. In this section, we first include reasoning theory from philosophy and logic and derive it into NLP reasoning. Then, we review some natural language reasoning topics in NLP. Finally, we propose a definition for reasoning in NLP, which combines the definition in philosophy and logic and the concerns of the NLP community.

\subsubsection{Definition from philosophy and logic} 

Here we introduce two descriptions and three definitions of reasoning from philosophy and logic: task-based description (Description \ref{desc:task}), negation-based description (Description~\ref{desc:neg}), logic-based definition (Definition~\ref{def:logic_reasoning}), assertion-based definition (Definition~\ref{def:theo_reasoning}), and action-based definition (Definition~\ref{def:prac_reasoning}). The former two descriptions can tell us ``what reasoning can do'' and ``what isn't reasoning'', while the latter three provide us different definitions of ``what is reasoning''. However, the definition from logic (Definition~\ref{def:logic_reasoning}) restricts reasoning to a subset within the coverage of formal logic. To reach a more generalized definition, we adopt the latter two definitions from philosophy, which are two different classes named theoretical reasoning and practical reasoning, respectively, as the basis for defining natural language reasoning in NLP.

\begin{desc}[task-based]\label{desc:task}
Reasoning is an essential mental activity when conducting conscious tasks with complex computations such as problem-solving, decision-making, persuasion, and explaining~\cite{thinking11,finocchiaro84,govier89,peter81}. 
\end{desc}

\begin{desc}[negation-based]\label{desc:neg}
Reasoning is a dynamic process to get some knowledge without direct recourse to sense perceptions or immediate experience, which is opposed to sensation, perception and feeling~\cite{peter81,walton90,Britannica-reason20}.
\end{desc}

\begin{definition}[logic-based reasoning]\label{def:logic_reasoning}
Reasoning is to discover valid conclusions by applying logic~\cite{locke1847,peter81,finocchiaro84,Britannica-reason20,walton90}.
\end{definition}

\begin{definition}[assertion-based reasoning / theoretical reasoning]\label{def:theo_reasoning}
Reasoning is to infer conclusions from a set of premises, consisting of one or more inference steps, where premises and conclusions are assertions that claim something is true or false about the world~\cite{Oxford08,peter81,walton90,Britannica-reason20,runes01}. 
\end{definition}

\begin{definition}[action-based reasoning / practical reasoning]\label{def:prac_reasoning}
Practical reasoning is to infer actions from goals and knowledge, which is oriented to deciding whether an action is practically reasonable~\cite{Oxford08,walton90}. 
\end{definition}

\subsubsection{Definition in NLP we suggest} 

According to Definition~\ref{def:theo_reasoning}, Definition~\ref{def:prac_reasoning} and negation-based description~\ref{desc:neg}, we can know ``what is reasoning'' and ``what isn't reasoning'' from the perspective of philosophy. There are also some descriptions towards the two questions in NLP. We compare and combine them in Table~\ref{tab:comp_def_reasoning}. We also review typical natural language reasoning datasets in NLP to observe and capture what the NLP community is concerned about. 

From our observations, in NLP, natural language reasoning also combines multiple knowledge to derive conclusions. The unique characteristics are (1) knowledge sources and (2) conclusion types. Firstly, common knowledge sources are knowledge bases, context, and PLMs, where the former two can explicitly provide encyclopedic knowledge and contextual knowledge, while the last is implicit knowledge sources. Secondly, In addition to assertions and actions, it is also popular to infer relations, e.g. causes and effects, of events. We demonstrate examples of these three conclusion types in Table~\ref{tab:conclusion}.

\begin{table}[ht]
    \centering\footnotesize
    \begin{tabular}{lll}
    \toprule
                            & \textbf{What is Reasoning}       & \textbf{What isn't Reasoning} \\
        \midrule
        \textbf{Philosophy}         & \makecell[l]{infer a new assertion from a set of assertions \\ infer an action from goals and knowledge}                    & \makecell[l]{sensation, perception and feeling\\direct recourse to sense perceptions or immediate experience} \\
        \hline
        \textbf{NLP}          & \makecell[l]{more than understanding, slow thinking \\ e.g. multi-hop QA, commonsense reasoning}          & \makecell[l]{memorize, look up, match information \\ e.g. text summarization, style transfer} \\
        \hline
        \textbf{Combination}   & \multicolumn{2}{l}{\makecell[l]{a dynamic process to integrate multiple knowledge to get new conclusions, \\rather than direct recourse to memorized or provided first-hand information}} \\
        \bottomrule
    \end{tabular}
    \caption{\label{tab:comp_def_reasoning}Comparison and combination of descriptions about reasoning from philosophy and NLP.}
\end{table}


\begin{table}[ht]
    \centering\small
    \begin{tabular}{llll}
    \toprule
                                 & \textbf{Premise}                                    & \textbf{Conclusion}    \\
        \midrule
        \textbf{Assertion}       & \makecell[l]{Cat is animal. \\Animal can breathe.}  & Cat can breathe.   \\
        \hline
        \textbf{Event}           & \makecell[l]{John was shot. \\There are people around.\\Doctor can save life.}                                      & John will be sent to see a doctor. \\
        \hline
        \textbf{Action}          & \makecell[l]{Marry is on the living room. \\Marry feels it hot. \\Remote control for air conditioner is in the bedroom.}    & \makecell[l]{go to the bedroom, take the remote control \\come back and turn on the air conditioner} \\
        \bottomrule
    \end{tabular}
    \caption{\label{tab:conclusion}Three types of conclusion in reasoning, where ``assertion'' and ``event'' assume something true or likely to be true in the world.}
\end{table}

Correspondingly, we propose the definition of NLP reasoning in Definition~\ref{def:NLP_reasoning} and suggest ``what isn't reasoning in NLP'' and ``what NLP reasoning can do'' in Description~\ref{desc:NLP_neg} and Description~\ref{desc:NLP_task}. It should be emphasized that conclusions are new (or unknown) assertions, events, or actions, which distinguishes reasoning from other knowledge-intensive tasks that may also require multiple knowledge. To better demonstrate the definition, we explain why some knowledge-intensive datasets are not reasoning in Table~\ref{tab:non_reason_dataset}.

\begin{definition}[NLP reasoning]\label{def:NLP_reasoning}
Natural language reasoning is a process to integrate multiple knowledge (e.g. encyclopedic knowledge and commonsense knowledge) to derive some new conclusions about the (realistic or hypothetical) world. Knowledge can be from both explicit and implicit sources.  Conclusions are assertions or events assumed to be true in the world, or practical actions.
\end{definition}

\begin{desc}[NLP negation-based]\label{desc:NLP_neg}
Natural language reasoning is to derive new assertions, events, or actions without direct recourse to models' memorization, knowledge base storage and the provided context.
\end{desc}

\begin{desc}[NLP task-based]\label{desc:NLP_task}
Reasoning is an important method to arrive at the required answers or solutions. It is effective when what we need is neither provided by context nor memorized by models and stored by knowledge bases, but reachable by integrating available information. 
\end{desc}

\begin{table}[ht]
    \centering\small
    \begin{tabular}{llll}
    \toprule
                                              & \textbf{Task}                   & \textbf{Why not reasoning}               \\
        \midrule
        \textbf{CoNLL}~\cite{AIDA-CoNLL11}     & entity linking     & \makecell[l]{just align known entities \\ without producing new assertions, events, or actions}   \\
        \hline
        \textbf{CommonGen}~\cite{CommonGen20}  & constrained text generation     & \makecell[l]{generate text \\ but neither true assertions or events, nor actions} \\
        \hline
        \textbf{Natural Questions}~\cite{NaturalQuestions19} & open-domain QA    & the answer can be simply matched \\
        \bottomrule
    \end{tabular}
    \caption{\label{tab:non_reason_dataset}Examples to explain what is not reasoning.}
\end{table}

\subsubsection{Key Concepts} 
We first introduce the key concepts: proposition and inference. Similarly, we derive the definitions from philosophy and logic to NLP. Then, we further clarify the definition of reasoning in NLP.

\paragraph*{Definition of key concepts} 
In logic, the proposition is the basic operation unit in reasoning, and inference is a sub-process of a complete reasoning process. Concretely, while reasoning is performed with statements (as premises and conclusions), the real operation units are the semantics behind sentences, i.e. propositions~\cite{hurley14}. Inference is a single step in reasoning~\cite{Oxford08,Britannica-infer17,walton90,runes01,Britannica-logic22}, and each reasoning can be made of one or more inference steps (Definition~\ref{def:theo_reasoning}. We put the two key concepts into NLP in Definition~\ref{def:nlp_proposition} and Definition~\ref{def:nlp_inference}.

\begin{definition}[NLP proposition]\label{def:nlp_proposition}
A proposition is the semantic meaning or information content of a statement rather than its superficial linguistic.
\end{definition}

\begin{definition}[NLP inference]\label{def:nlp_inference}
Inference is a single step that produces a single (intermediate) conclusion from some premises.
\end{definition}

\begin{figure}[ht]
\centering\scriptsize
\includegraphics[width=0.5\linewidth]{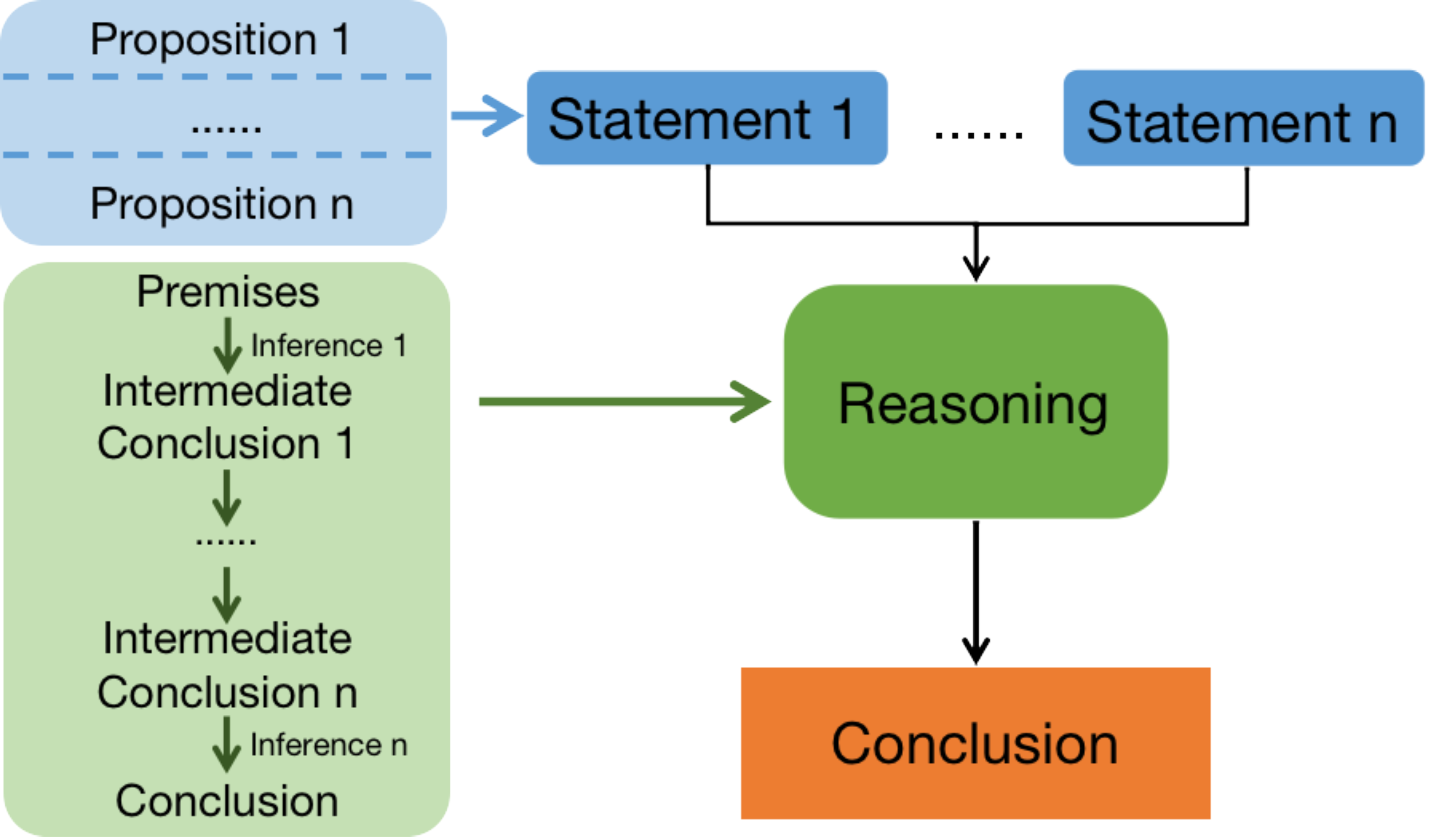}
\caption{Reasoning process. The premises can be either explicit or implicit knowledge, e.g. PLMs' memory.}
\label{fig:reasoning_process}
\end{figure}

\begin{figure}[ht]
\centering\scriptsize
\includegraphics[width=0.9\linewidth]{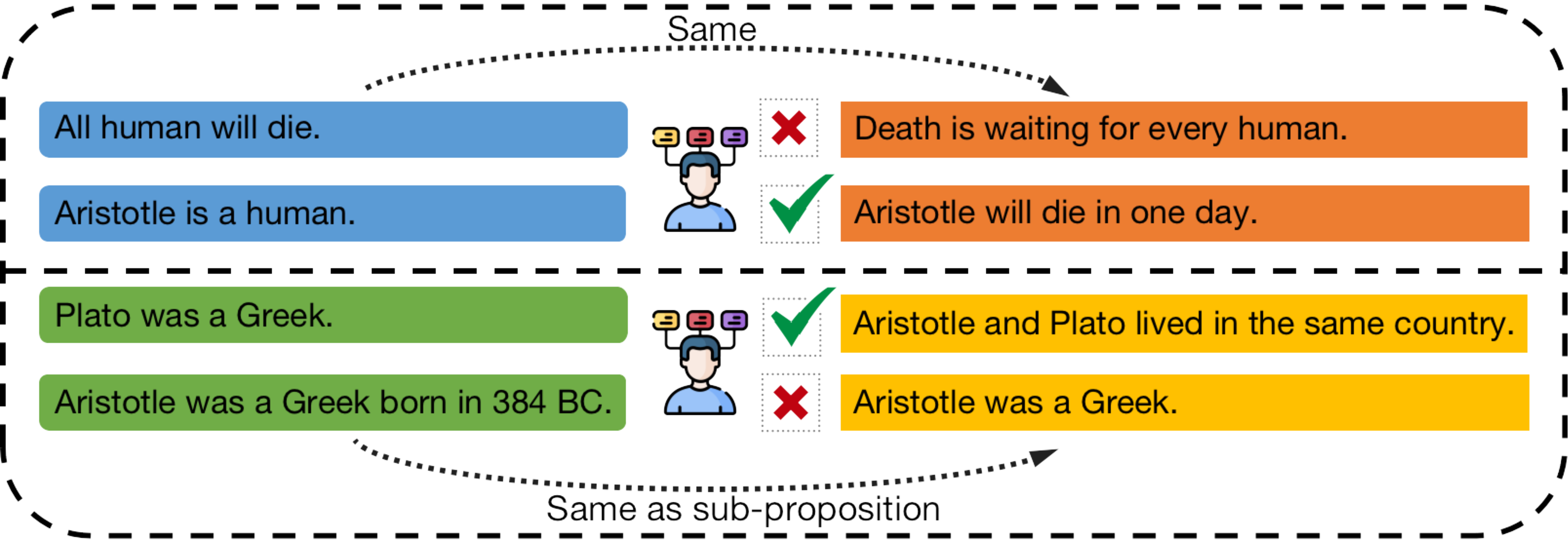}
\caption{Examples to show the key idea of ``semantic difference'', where check mark denotes reasoning while cross denotes not reasoning.}
\label{fig:reasoning_examples}
\end{figure}

\paragraph*{Further clarify the definition of NLP reasoning} 
We leverage these concepts to clarify the definition of NLP reasoning: what we mean by ``integrate multiple information to derive new conclusions'' is that (1) a single sentence conveying multiple semantics can provide multiple premises, (2) there must yield new semantics in inference and reasoning, i.e. conclusions are semantically different to all premises. We detail two examples to demonstrate this key idea (Fig~\ref{fig:reasoning_examples}) and illustrate the definition of reasoning in Fig~\ref{fig:reasoning_process}.

\subsection{Categories of Inference}\label{sec:category}
While knowledge has been well categorised in NLP (e.g. explicit world knowledge and implicit commonsense knowledge), we find that there is still a lack of reasonable taxonomy for inference. Therefore, we borrow the categories from philosophy and discuss the differences between classes to NLP and how they can affect the solutions.

Inference can be divided into (mainly) deductive, inductive and abductive~\cite{peirce92,peirc-IEP}, or divided into monotonic and defeasible. Actually, the deduction is monotonic inference while induction and abduction are sub-classes of defeasible inference. Since ``monotonic'' and ``defeasible'' can capture the difference between deductive and non-deductive inference, we combine the two taxonomies into one: deductive inference and defeasible inference.

\subsubsection{Deduction, Induction, and Abduction} 
According to Aristotle and Peirce, there are three major inferences: deduction, induction, and abduction~\cite{peirce92,peirc-IEP}. This taxonomy is the most familiar one to the NLP community, adopted and studied by several works~\cite{bAbI16,ruletaker20,LGID22,FOLIO22,ART20,AbductionRules22,CLUTRR19,DEER22}. The definitions are shown below (Table~\ref{tab:ex_diff_dia} shows examples).

\begin{definition}[Deduction]\label{def:deduction}
A deductive inference is to infer valid knowledge (conclusion) from the given knowledge (premises).
\end{definition}

\begin{definition}[Induction]\label{def:induction}
An inductive inference is to infer probable knowledge, which describes a more general rule, extrapolated from the given knowledge.
\end{definition}

\begin{definition}[Abduction]\label{def:abduction}
An abductive inference is to infer probable knowledge, as the best explanation (i.e. cause), for the given knowledge (i.e. phenomena). 
\end{definition}

\begin{table}[ht]
    \centering
    \begin{tabular}{c|c|c}
    \toprule
        \multicolumn{3}{l}{Fact1: Aristotle is a human} \\
        \multicolumn{3}{l}{Rule: All human will die} \\
        \multicolumn{3}{l}{Fact2: Aristotle will die} \\
        \hline
        \textbf{Deduction} & \textbf{Abduction} & \textbf{Induction}\\
        (Fact1 + Rule $\rightarrow$ \red{Fact2}) & (Fact1 + \red{Rule} $\leftarrow$ Fact2) & (Fact1 + Fact2 $\rightarrow$ \red{Rule})\\
        \bottomrule
    \end{tabular}
    \caption{\label{tab:ex_diff_dia}An simple example to show the difference between deduction, abduction and induction, where text in black is the given knowledge while text in red is the inferred knowledge. ``Fact'' denotes specific knowledge while ``rule'' denotes general principle.}
\end{table}

However, among these three classes, researches on abduction and induction are much under-explored than deduction, while the widely studied deduction is only a very small set of human daily reasoning.

\subsubsection{Deductive Inference and Defeasible Inference} 
Our main goal is to promote research on non-deductive reasoning and highlight the differences and challenges. Therefore, we turn into monotonic inference and defeasible inference, which can better capture the features of deductive and non-deductive inference respectively. 

\paragraph*{Key difference} 
The key difference between monotonic inference and defeasible inference from philosophy is that the former derives valid conclusions\footnote{``Valid'' means when the premises are true, the conclusion is impossible to be false.} while the latter only produces probable conclusions. Since the conclusions of deductive inference are truth-preserving that the future added knowledge will not affect their validity, thus the set of knowledge is incremental, i.e. monotonic. By contrast, the conclusions of non-deductive inference (e.g. induction and abduction) may be wrong, and the newly added knowledge may retract the conclusion, i.e. defeasible. For example, one may inductively infer ``birds can fly'' with the premises ``parrots can fly'' and ``eagles can fly''. However, when he or she discovers the new knowledge ``ostrich cannot fly'', the conclusion will be retracted.

\paragraph*{Different characteristics} 
This difference towards conclusions between deductive inference and defeasible inference leads to many different characteristics, including inference relations between premises and conclusions, the quality of inference, and the requirement of knowledge. Concretely, there is only one inference relation between premises and each conclusion in deductive inference, i.e. support, and the inference is either valid or invalid. Therefore, we can derive a valid conclusion just with several supporting premises. By contrast, knowledge can strengthen, weaken and even rebut (the probability of) the conclusion in defeasible inference, and the quality of inference varies from weak to strong. Therefore, it is better to collect more comprehensive information to arrive at a more probable conclusion. We compare the characteristics of the deductive inference and defeasible inference in Table~\ref{tab:comp_inference}.

\begin{table}[ht]
    \centering
    \begin{tabular}{lll}
    \toprule
                                        & \textbf{Deductive Inference}       & \textbf{Defeasible Inference} \\
        \midrule
        \textbf{Conclusion}             & true                               & probably true \\
        \hline
        \textbf{Inference relation}     & support                            & strengthen, weaken, rebut \\
        \hline
        \textbf{Quality of inference}   & valid or invalid                   & weak to strong \\
        \hline
        \textbf{Required knowledge}     & bounded                            & unbounded \\
        \bottomrule
    \end{tabular}
    \caption{\label{tab:comp_inference}The characteristics of the deductive inference and defeasible inference.}
\end{table}

\paragraph*{Affects on NLP} 
These characteristics affect relevant knowledge acquisition, reasoning path structure, and the importance of interpretability in NLP. Firstly, while collecting the supporting knowledge toward the valid conclusion is enough for deductive reasoning, it is better to collect both supportive and opposing knowledge to compare the confidence of different conclusions for defeasible reasoning. Then, there has been increasing attention on reasoning path generation in NLP~\cite{PRover20,COT22,Entailer22}. However, due to more types of inference relation, the structure of reasoning paths for defeasible reasoning is more complex than deductive reasoning and thus becomes more challenging to generate. Finally, it is more important and sometimes even crucial for  NLP models to perform interpretable defeasible reasoning. This is because people with different background knowledge can infer very different and even opposite conclusions by themselves, thus it is much more difficult to clarify the conclusion without explicit premises and reasoning procedure.

\subsection{Potentials, Challenges, and Requirements of NLR}\label{sec:nlr_pc}

\paragraph*{Potentials} 
Compared to reasoning with precise formal language, natural language provides a better human-computer interaction interface. Besides, natural language opens a door to play with defeasible reasoning, where formal language fails. 

\paragraph*{Challenges} 
Firstly, natural language suffers from ambiguity and variety, since there are polysemy, synonymy and diverse structures. Therefore, while triples and formal languages are precise, statements and propositions are many-to-many in natural language, which poses a challenge on natural language understanding. Secondly, supervised data of inference is difficult to obtain, which may prevent it from large-scale training. Moreover, the step of reasoning is diverse at the instance level, i.e. different questions may require different inference steps to answer, and it is important to generalize to the unseen steps. 

\paragraph*{Requirements} 
Based on the definition (Definition~\ref{def:NLP_reasoning}), the key components in NLP to achieve reasoning are (1) (multiple) knowledge and (2) an algorithm capable of understanding and inference. Correspondingly, there are three stages: knowledge acquisition, knowledge understanding, and inference. Firstly, it requires collecting the relevant knowledge required for reasoning (knowledge acquisition). Then, the algorithm requires to capture propositions underlying the given knowledge (knowledge understanding). In addition to the general semantics, it should also capture the logical semantics such as negation, conjunction and disjunction. Subsequently, beginning from these propositions, the algorithm requires integrating some knowledge to infer a new conclusion with one or more steps to reach the final answer (inference). Though knowledge acquisition and understanding are also necessary for reasoning, the two topics are big enough to write another survey, thus we just focus on inference in this article.

\section{Why PLMs for Natural Language Reasoning}\label{sec:why}

\subsection{Introduction to PLMs}
Pre-trained language models (PLMs) are based on transformer architecture~\cite{transformer17}, which is built with many attention modules and are pre-trained on massive amounts of text data via unsupervised learning techniques such as predicting masked tokens~\cite{BERT19} or generating the next tokens~\cite{GPT-18}. Since BERT~\cite{BERT19} occurred, pretraining-then-finetuning became a common paradigm, which transfers the general abilities of PLMs learned in the pretraining stage to downstream tasks with further task-specific finetuning. Since large language models have been found to be few-shot learners~\cite{GPT3-20}, in-context learning has become a new popular paradigm, which can predict a new sample with only a few demonstrations without finetuning parameters. Recently, the zero-shot prompting paradigm also becomes more popular in LLMs~\cite{LLM-ZSR22}.

\paragraph*{Types of PLMs} 
According to the architecture, PLMs can be divided into encoder-only (e.g. BERT~\cite{BERT19}), decoder-only (e.g. GPT~\cite{GPT-18}) and encoder-decoder (e.g. T5~\cite{T5-19}). According to the directivity, PLMs can be divided into bidirectional (encoder-only) and causal (decoder-only and encoder-decoder), while bidirectional PLMs are commonly used for discriminative tasks, causal PLMs can model general tasks but are more capable of generative tasks. According to the model size, there are medium-size PLMs and large language models, where LLMs are much larger than the former (e.g. 13B parameters).

\paragraph*{Advantages of PLMs for NLR} 
We conclude with four advantages of PLMs for NLR. 
\begin{itemize}
    \item \textbf{Ability of natural language understanding}. Transformers represent words and sentences in a context-dependent manner as continuous vectors in a high-dimensional space dealing with ambiguity and uncertainty in nature. After large-scale pretraining, PLMs can learn a powerful understanding capability, which helps them to capture and understand knowledge mentioned in the text. 
    
    
    \item \textbf{Ability to learn implicit knowledge into parameters}. It has been found that PLMs can capture some implicit knowledge that is not explicitly mentioned, such as commonsense knowledge, into their parameters. This is important since it is impossible to explicitly enumerate and provide commonsense knowledge for reasoning. 
    
    \item \textbf{Ability of in-context learning}. LLMs such as GPT-3 exhibit the impressive ability to perform tasks only with some demonstrations without further fine-tuning, which is valuable to alleviate data sparsity problems. 
    
    \item \textbf{Emergent abilities}. Recently, it was found that LLMs have some emergent abilities that only occur when the model size is big enough~\cite{emer22}, and LLMs can perform much more complex tasks as their size increases. Moreover, it has been demonstrated that performing multi-step reasoning in a few-shot or zero-shot manner is one of the emergent abilities~\cite{COT22}. 
\end{itemize}


\subsection{Empirical Development}

Recent progresses also show the potential to leverage PLMs on natural language reasoning, which exhibits their learning and generalization abilities of reasoning skills with both explicit and implicit knowledge. 

By finetuning on the specific dataset, ~\cite{ruletaker20} first demonstrated that PLMs can perform deductive reasoning over explicitly provided natural language statements, which can zero-shot transfer to different domains. Moreover, ~\cite{LeapOfThoughtT20} showed that PLMs can combine memorized implicit taxonomic and world knowledge with explicitly provided knowledge for the deduction. In addition to deduction, PLMs can also learn to perform defeasible reasoning~\cite{defeasibleNLI20,AbductionRules22,DEER22}.

While LLMs with in-context learning were once thought to be incapable of multi-step reasoning, it has been found that their capabilities of reasoning can be unlocked by generating forward reasoning paths before the final answer~\cite{COT22}, which is called Chain-of-Thought (CoT) prompting. With this prompting, the performance of many multi-step reasoning tasks in Big-Bench Hard can surpass the average human rater. Furthermore, LLMs can perform multi-step reasoning not only with few-shot exemplars, ~\cite{LLM-ZSR22} also found that they can automatically produce intermediate steps with a simple ``Let's think step by step'' prompting in a zero-shot manner. Surprisingly, LLMs can even learn from their self-generated reasoning paths~\cite{STaR22,self-improve22}. Moreover, GP4 outperformed a majority of people on several realistic examinations such as Uniform Bar Exam which also require some reasoning.

In addition, to forward reasoning paths, question decomposition, a backward reasoning method, is also effective in multi-hop question answering, which is beneficial to both medium-size PLMs~\cite{DecompRC19,QD22} and LLMs~\cite{QD22,self-ask22}.

Moreover, while neural-based methods are blamed for black box prediction, ~\cite{FaiRR22,faithful22} demonstrated that PLMs can produce faithful reasoning paths and make predictions based on them.

In conclusion, PLMs can learn to perform multi-step reasoning from supervised data or few-shot demonstrations. Their capabilities of natural language understanding, generalization, and leveraging implicit knowledge make them promising to deal with arbitrary natural language, commonsense knowledge and defeasible reasoning.

\section{Methodologies of NLR}\label{sec:method}

\begin{figure*}[tp]
    \centering
    \begin{forest}
        forked edges,  
        for tree={
            grow=east,
            anchor=base west,  
            reversed=true,
            font=\small,
            align=left,
            draw=black,  
            rounded corners,  
            minimum width=4em,
            s sep=3pt,
            inner xsep=2pt,  
            inner ysep=3pt,  
        },
        where level=1{font=\scriptsize,}{},
        where level=2{font=\scriptsize,}{},
        where level=3{font=\scriptsize,}{},
        [
            Natural Language Reasoning
            [
                End-to-End Reasoning \\ (\S~\ref{sec:end-to-end})
                [specialized models]
                [vanilla medium-size PLMs]
                [vanilla decoder-only LLMs]
                [specialized pretraining]
            ]
            [
                Forward Reasoning \\ (\S~\ref{sec:forward})
            ]
            [
                Backward Reasoning \\ (\S~\ref{sec:backward})
                [backward chaining]
                [question decomposition]
            ]
        ]
    \end{forest}
    \caption{Taxonomy of natural language reasoning}
    \label{fig:taxonomy_of_nlr}
\end{figure*}
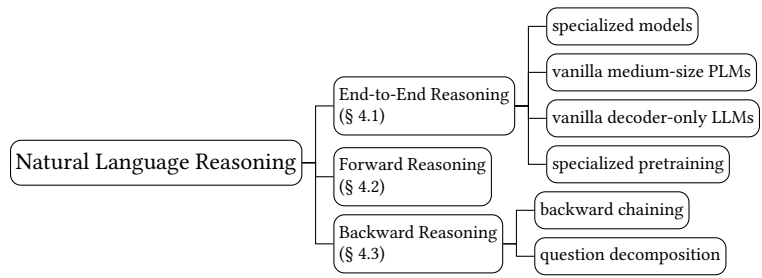

In this section, we introduce three types of natural language reasoning approaches: end-to-end reasoning (Sec~\ref{sec:end-to-end}), forward reasoning, and backward reasoning. The overall taxonomy is shown as Figure~\ref{fig:taxonomy_of_nlr}.

The key difference among these three categories lies in the reasoning path. Concretely, ``end-to-end reasoning'' only predicts the final answers without any intermediate text, while the latter two approaches can produce reasoning paths, containing one or more steps with the intermediate conclusions, showing the process of (possibly multi-step) reasoning that links premises to the conclusion\footnote{There are also some researches on producing natural language explanations instead of reasoning procedure, but we just focus on reasoning paths in this survey}. 

Presenting the reasoning path for each prediction can improve the interpretability of a system. Especially, a strict reasoning path can also explicitly expose the supporting knowledge of each step. Moreover, producing reasoning paths has been demonstrated to be beneficial to the final performance of multi-step reasoning~\cite{COT22,LLM-ZSR22,BBH22,QD22,self-ask22}. There are two directions of reasoning.

\paragraph*{Two Directions of reasoning} 
Multi-step reasoning can be performed by either forward~\cite{ProofWriter21,FaiRR22,COT22,selection-inference22} or backward~\cite{DecompRC19,EVR21,modularqa21,Entailer22,self-ask22}. Forward reasoning is a bottom-up procedure, which starts from the existing knowledge and repeatedly makes inferences to obtain new knowledge until the problem is solved. The other, backward reasoning, is a top-down procedure, which starts from the problem and repeatedly breaks down into sub-problems until all of them can be solved by the existing knowledge. While backward reasoning targets the specified problems, forward reasoning can freely uncover new knowledge implicated by the existing knowledge without preassigned problems. Accordingly, the search space of forward reasoning is much larger than backward reasoning when solving a specific problem, facing the combinatorial explosion as the step of inference goes. When it comes to theorem proving, which is a verification problem, where the reasoning path is named ``proof'', forward reasoning and backward reasoning are often called ``forward chaining'' and ``backward chaining'' respectively.

\begin{figure}[ht]
\centering\scriptsize
\includegraphics[width=0.92\linewidth]{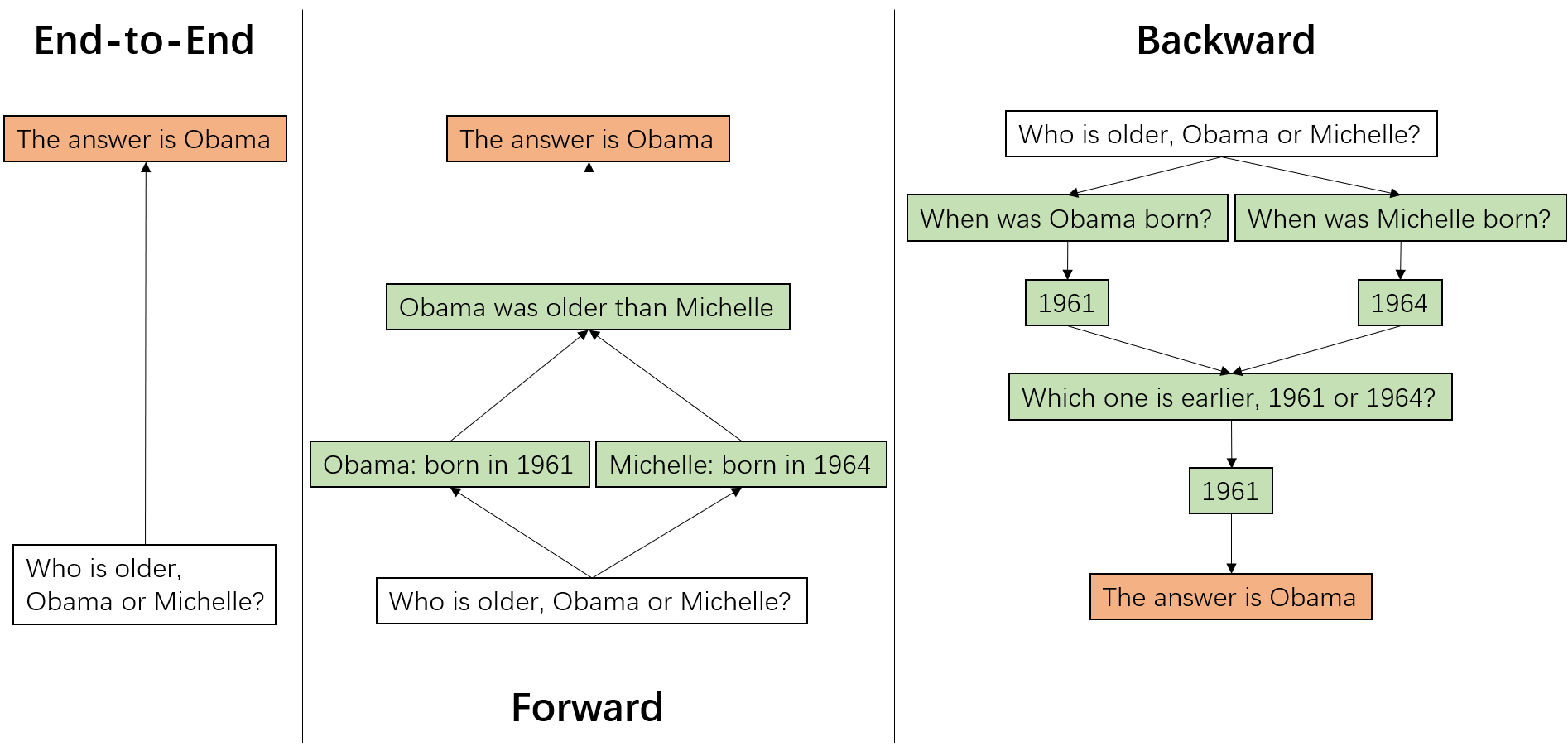}
\caption{An example to demonstrate the reasoning procedure of end-to-end reasoning, forward reasoning, and backward reasoning. White colours the question, green colours the intermediate text, and orange colours the answer.}
\label{fig:reasoning_example}
\end{figure}

We compare these three methods in Table~\ref{tab:comp_reasoning} and demonstrate an example in Figure~\ref{fig:reasoning_example}. The following subsections will further introduce and discuss the comparison.

\begin{table}[ht]
    \centering\normalsize
    \begin{tabular}{llll}
        \toprule
                              & \textbf{Direction}   &  \textbf{Pros}   &  \textbf{Cons}  \\
        \midrule
        \textbf{End-to-End Reasoning} & -            & most efficient &  \makecell[l]{blackbox \\ bad generalization} \\
        \hline
        \textbf{Forward Reasoning}    & bottom-up    & \makecell[l]{interpretability \\ open-ended} & \makecell[l]{huge search space \\ only effective in LLMs}  \\
        \hline
        \textbf{Backward Reasoning}   & top-down     & \makecell[l]{interpretability \\ efficient} & goal-specific  \\
        \bottomrule
    \end{tabular}
    \caption{\label{tab:comp_reasoning}Comparison of end-to-end reasoning, forward reasoning, and backward reasoning.}
\end{table}

\subsection{End-to-End Reasoning}\label{sec:end-to-end}
End-to-end reasoning is a complete black-box prediction that only outputs the final answers without any explanation, intermediate conclusion, or reasoning path, whether it is a single-step or multi-step reasoning problem. There are mainly three kinds of models used to perform end-to-end reasoning: specialized models built upon medium-size PLMs, vanilla medium-size PLMs, and decoder-only LLMs. Besides, there is also some research on specialized pretraining methods.

\subsubsection{Training specialized models} 

To perform end-to-end reasoning, models need to aggregate multiple knowledge and reason over them. Correspondingly, there are specialized models improving the capability of multiple evidence aggregation~\cite{Transformer-XH20,ege21,thinkaboutit21,GreaseLM22} or reasoning~\cite{E3-19,OpenCSR21,DFGN19,MHGRN20,QA-GNN21}. Previous research often incorporated some task-specific inductive biases via architectural designs. For example, graph neural networks are popularly used to leverage edges (e.g. entity-entity relations) to promote information aggregation and integration between nodes (e.g. entity information)~\cite{DFGN19,MHGRN20,QA-GNN21}. However, these designs only specialize in either specific tasks or datasets. By contrast, ReasonFormer~\cite{ReasonFormer22} proposed a variant architecture of transformer for general reasoning, with different modules responsible for different predefined fundamental reasoning capabilities. This kind of model can improve performance on specific tasks or datasets. Nevertheless, all of these designs rely heavily on handcrafts, introducing strong prior assumptions, which may hurt the generalization ability to other tasks.

\subsubsection{Finetuning vanilla medium-size PLMs} 

Medium-size PLMs lack the ability to perform zero-shot reasoning such as theorem proving, argument completion, commonsense reasoning, and abduction without training~\cite{ruletaker20,AAC21,RICA21,AbductionRules22}. Recently, it was found that transformers can be good soft deductive reasoners after in-domain training~\cite{ruletaker20,AAC21}. By contrast, it is more challenging to perform defeasible reasoning~\cite{defeasibleNLI20,AbductionRules22,DEER22}. 

\paragraph*{Deductive reasoning} 
Both bidirectional and causal PLMs have demonstrated learning ability for deductive reasoning. ~\cite{ruletaker20} first found that BERT and RoBERTa (bidirectional PLMs) can perform theorem proving over synthetic natural language facts and rules after training. When it comes to causal PLMs, ~\cite{AAC21} demonstrated that GPT2 can learn to reason over deductively valid arguments and is able to generalize from simple core schemes to some unseen composite schemes. However, there are two challenging problems in this paradigm: data sparsity and spurious correlations. 

Due to data sparsity, many researchers resort to synthetic data, which is far away from the realistic setting~\cite{ruletaker20,AAC21,ProofWriter21}. Moreover, researchers demonstrated that training RoBERTa on synthetic data fails to generalize to linguistic variations on theorem proving and commonsense reasoning~\cite{ruletaker20,RICA21}, which indicates they learn less of the general logical structure underlying the linguistic variations. While training on high-quality data~\cite{FOLIO22} can alleviate the spurious correlation problem~\cite{FOLIO22}, such data is difficult to annotate on a large scale. Although automatic data collection can obtain large-scale examples, it is restricted to limit reasoning types dependent on the designed heuristic methods~\cite{ParaPattern21}.

On the other hand, PLMs are found to learn spurious correlations on multi-hop reasoning, theorem proving and commonsense reasoning~\cite{single-hop-rc19,paradox22,RICA21}. In other words, finetuning on specific tasks and datasets may lead models to overfit to the specific spurious correlations underlying them. There are several researchers trying to reduce artifacts in the dataset such as by adding adversarial data~\cite{Adversarial-MultiHopQA19} and carefully constructing the new dataset~\cite{2wikimultihop20,musique22}. However, it is difficult to construct data without any artifact, and there may be some statistical features inherent in the problem which cannot avoid in principle~\cite{paradox22}. Another line to alleviate shortcuts is increasing attention on reasoning path generation, which may encourage models to perform actual reasoning (Sec~\ref{sec:forward}).

\paragraph*{Defeasible reasoning} 
The capability of defeasible reasoning seems to be more challenging for vanilla medium-size PLMs to learn. Specifically, ~\cite{defeasibleNLI20} demonstrated that the performance of BART-large and T5-large on a defeasible reasoning task, i.e. generate a statement to update the strength of a probable conclusion, is far from satisfaction. There is a similar observation on inductive reasoning~\cite{DEER22}. Besides, it is hard to generalize the ability of abduction learned in a synthetic dataset to unseen domains~\cite{AbductionRules22}\footnote{By contrast, the ability of deduction learned in the synthetic dataset can be generalized to other domains~\cite{ruletaker20}}. While data sparsity is also a challenge for defeasible reasoning, how to better enable PLMs capable of defeasible reasoning remains a problem.

\subsubsection{Few-shot decoder-only LLMs} 

Few-shot prompting using decoder-only LLMs without finetuning can alleviate data sparsity and also prevents models from overfitting to specific tasks or datasets. However, there remains the question of whether models can be better capable of reasoning as the model size increases.

Although the performance on reasoning problems improves as the model size increases~\cite{selection-inference22,FOLIO22,defeasibleNLI20,DEER22}, it is still unclear how much progress can be attributed to the improvement in reasoning capability. ~\cite{selection-inference22} demonstrated that (deductive) reasoning problems are much more challenging that the scaling laws (of the Gopher family) work much slower than other tasks in BigBench and vanilla LLMs struggle with multi-step reasoning problems. ~\cite{self-ask22} found that while LLMs memorize more factual knowledge as the model size increases, it seems their ability to implicitly integrate knowledge for deduction does not improve. 

Surprisingly, more reasoning capabilities of LLMs can be elicited by chain-of-thought prompting, as introduced in Sec~\ref{sec:forward}.

\subsubsection{Specialized pretraining}
To improve the reasoning capability of PLMs, there is some research on introducing inductive biases of reasoning when continual pretraining~\cite{ReasonBERT21,ContextualSP22,MERIt22,APOLLO22}. There are type-specific inductive biases~\cite{MERIt22,ContextualSP22} and type-agnostic inductive biases~\cite{ReasonBERT21,APOLLO22}. For example,~\cite{ReasonBERT21} incorporated the general inductive bias of reasoning over multiple long evidence texts, while~\cite{MERIt22} mainly designed for relational reasoning. Inductive biases are introduced with reasoning-related data and training strategies. For example,~\cite{APOLLO22} collected reasoning-related text that involves logical inference keywords and let models to self-supervised predict these keywords. When the pretraining improves performance on multi-hop reasoning and logical reasoning problems, especially in the low-resource setting~\cite{MERIt22,ReasonBERT21,APOLLO22}, all of them worked on encoder-only PLMs, i.e. BERT and RoBERT. 

Recently,~\cite{ContextualSP22} proposed a new line that leverages programs such as SQL to pretrain PLMs with synthesized (program, execution result) pairs. The results are inspiring that PLMs, including medium-size, large-size, encoder-only and encoder-decoder, can attain significant improvement in multi-hop reasoning and logical reasoning. 

However, it is important to ask whether it is still beneficial to incorporate inductive biases into LLMs, or whether simply increasing the model size and pretraining on more data is enough to improve reasoning capability. In other words, can LLMs learn reasoning well enough just by the current general pretraining? Maybe LLMs have already learned powerful reasoning capabilities that just need to be elicited via smart prompting such as CoT~\cite{COT22}.

\subsection{Forward Reasoning}\label{sec:forward}
Forward reasoning repeatedly composes the existing knowledge to derive new knowledge until reaching the answers. There are two kinds of benefits to producing a forward reasoning path: trustworthiness~\cite{EntailmentBank21,FaiRR22,faithful22} and performance improvement~\cite{COT22,selection-inference22,BBH22}. 

\subsubsection{Trustworthiness} 

Showing how multiple knowledge interacts and contribute to new conclusions can contribute to the system's interpretability. Furthermore, when the prediction is based on the reasoning procedure, it can alleviate the widespread shortcut problem. To exhibit the structure of reasoning, involving the required knowledge and their inference relation, reasoning paths are often represented as directed graphs or trees~\cite{PRover20,EntailmentBank21,MetaLogic22,defeasiblegraphs21}. Typically, each node represents one piece of knowledge and the edge represents the inference relation between knowledge. For example, a single inference linking two premises to one conclusion can be represented as two nodes linking to their shared parent node. 

\paragraph*{Deductive reasoning} 
There is only one inference relation in deductive reasoning, i.e. support. To construct such an interpretable reasoning path, it needs to find the relevant knowledge as premises and infer the conclusions (inference). Since inference is to produce new knowledge with the given premises, it is usually implemented by vanilla generative PLMs~\cite{FaiRR22,IRGR22}. Instead of explicitly selecting or retrieving the required knowledge~\cite{PRover20}, some works put the context into the input and modelled both knowledge selection and inference as unified generation~\cite{ProofWriter21,NLProofS22}. However, it may generate hallucinations and invalid inferences. To alleviate this problem, ~\cite{NLProofS22} leveraged an additional verifier to score the validity. In addition to just improving the validity of the knowledge node and inference relation edge, some researchers proposed performing faithful reasoning, which forces the prediction to rely on reasoning paths. This is mainly realized by designing decoupled modular frameworks to avoid shortcuts to irrelevant context~\cite{FaiRR22,METGEN22,faithful22}. For example, ~\cite{faithful22} iteratively performed knowledge selection and inference alternately in a step-by-step manner, where each inference step only conditions the currently selected knowledge to infer the conclusion without seeing the question and the previous steps. Both supervised modular frameworks based on medium-size PLMs~\cite{FaiRR22,METGEN22} and in-context learning modular frameworks based on LLMs~\cite{faithful22} have been explored to perform faithful reasoning. In addition to faithfulness, such step-decoupling behaviors also bring other effects. On the one hand, it is easier to provide supervised training data or in-context exemplars. The supervised framework can leverage one-step supervision~\cite{FaiRR22,METGEN22} to train the system, which alleviates the data sparse problem in multi-step reasoning, while the in-context learning framework can demonstrate representative one-step examples that avoid the challenge of selecting the appropriate exemplars for multi-step reasoning~\cite{selection-inference22,faithful22}. On the other hand, it brings error propagation. However, all of these works consider the simplest setting, where all the required knowledge is either explicitly provided in context or retrievable from knowledge bases.

\paragraph*{Defeasible reasoning} 
There are more types of inference relations in defeasible reasoning, i.e. strengthen, weaken (the probability of the conclusion) and rebut. Since it is difficult to collect all the supporting premises, researches on this line mainly concern the label of inference relations between statements. In other words, there exists implicit reasoning, i.e. some premises are not explicitly provided. Similar to deductive reasoning, reasoning paths can be generated by one-shot generation~\cite{MetaLogic22,defeasiblegraphs21} or faithful modular framework~\cite{MetaLogic22}. Different to deductive reasoning, it is more challenging to generate defeasible reasoning paths that even finetuned LLMs (T5-11B) find difficult.

However, there remains a problem with the evaluation of the constructed reasoning path. Specifically, there may be multiple reasoning paths for each problem, which poses challenges on data annotation~\cite{EntailmentBank21} and automatic evaluation~\cite{faithful22}. Annotating all possible reasoning paths for evaluation is impractical, especially for those long-step problems facing combinatorial explosion. And it is also challenging to automatically evaluate the validity of reasoning paths without annotated data.


\subsubsection{Performance improvement} 
Reasoning path can also be used to improve the answer performance on multi-step deductive reasoning, including the in-domain performance of LLMs and the generalization ability of PLMs. For this purpose, it is not necessary to involve all the required knowledge in the reasoning path or keep the validity of inferences as what we are concerned about are the final results rather than reasoning paths. 

Firstly, reasoning paths can improve the in-domain performance by providing enriching context~\cite{COT22,BBH22} or supervision signal~\cite{Flan22,ALERT22}. Recently, ~\cite{COT22} demonstrated that the LLMs' performance of several reasoning tasks such as commonsense reasoning (both deductive and defeasible) can be significantly improved by generating a reasoning path before the final answers, which is called chain-of-thought prompting (CoT). Before this, while LLMs are successful in classical NLP tasks, they fail in reasoning, especially multi-step reasoning tasks. This finding boosted a series of research on this line~\cite{LLM-ZSR22,Auto-CoT22,self-consistency22,selection-inference22,faithful22,BBH22,ALERT22,PRONTOQA22,MMRBES22,PDPS22,TSLMR22,reasoning-teacher22}. Especially, ~\cite{LLM-ZSR22} showed that even a simple zero-shot prompting ``let's think step by step'' can activate LLMs to perform commonsense reasoning and attained impressive performance. Furthermore, ~\cite{self-consistency22} found that the final performance on commonsense reasoning can be greatly further improved by just voting the results on multiple reasoning paths. Besides, in addition to performing reasoning on downstream tasks via few-shot prompting without changing the parameters, supervised finetuning LLMs on CoT annotations can further improve their reasoning capability~\cite{Flan22,ALERT22}. In addition to commonsense reasoning, the performance of classical logical reasoning and multi-step reasoning are also improved significantly by generating CoT~\cite{selection-inference22,BBH22}. However, classical logical reasoning is much more challenging than other typical tasks~\cite{selection-inference22}. Instead of one-shot CoT generation,~\cite{selection-inference22} proposed a more inspiring framework (SI) for theorem proving (a task of classical deductive reasoning) based on modules with different prompting, which outperforms 40x larger LLMs with CoT. Moreover, ~\cite{STaR22,self-improve22} found that LLMs can self-improve their reasoning capabilities by finetuning their self-generated reasoning paths. However, such abilities are only effective in LLMs, i.e. the model scale should be large enough, which is also seen as an emergent ability of LLMs that can be elicited by few-shot~\cite{COT22} and even zero-shot prompting~\cite{LLM-ZSR22}. There are some researches transferring the CoT reasoning capability of LLMs to smaller models via knowledge distillation~\cite{PDPS22,TSLMR22,reasoning-teacher22}. 

Moreover, it can improve the generalization ability of PLMs. It has been observed that constructing the proof graph for the goal hypothesis can improve the zero-shot generalization ability of medium-size PLMs to the unseen step of reasoning~\cite{PRover20,multiPRover21,ProofWriter21} and to unseen domain~\cite{ProofWriter21,METGEN22} on the theorem proving task, which is likely because it forces models to perform reasoning rather than exploit shortcuts. Also, turning one-shot construction into a stepwise procedure has a better generalization to the unseen steps of reasoning and to cross datasets~\cite{ProofWriter21,METGEN22}. 

However, the search space of forward reasoning suffers from the combinatorial explosion as the number of reasoning steps increases. In addition to performing a single-step inference, planning is also very important to multi-step reasoning, especially to deep steps. It has been observed that while LLMs are capable of a single inference, they still struggle to plan on deep reasoning steps~\cite{selection-inference22,PRONTOQA22}. Yet this topic is under-explored with few researches~\cite{faithful22,NLProofS22}.

In addition to deductive reasoning, leveraging reasoning paths to improve performance on defeasible reasoning is still under-explored.

\subsection{Backward Reasoning}\label{sec:backward}
Backward reasoning repeatedly breaks down problems into sub-problems and solves them until reaching the answers. Similar to forward reasoning, it can be used to produce trustworthy reasoning paths explicitly represented with knowledge and inference relations~\cite{IBR22,METGEN22,Entailer22} or improve the final performance without strict structures~\cite{DecompRC19,modularqa21,LAMBADA22}. It faces a smaller search space and thus is more efficient than forward reasoning. There are two popular backward reasoning methods: backward chaining and question decomposition. While the former is a proof-finding strategy, the latter is a general strategy available for general problems. Researches mentioned in this section are mainly about deductive reasoning.

\subsubsection{Backward Chaining} 
Backward chaining is the preferable approach for proof-finding by humans. Beginning from the goal, it repeatedly performs abductive reasoning to derive the potential premises as sub-goals until all the sub-goals can be proved or disproved by the existing knowledge. According to the source of the premises, or sub-goals, there are two kinds of abduction: predict part of premises (others are the existing knowledge) and predict all premises. The first one is to predict the unknown required premise for a conclusion with the existing knowledge, which can be realized by vanilla generative PLMs, either medium-size~\cite{METGEN22} or large size~\cite{LAMBADA22}. The other one is to predict all the premises from scratch without relying on the existing explicit knowledge, which can be realized by LLMs~\cite{Entailer22}. While the former kind of abduction is easier to perform, the latter can solve the scenario where all the required premises do not exist in the knowledge base. Compared to forward chaining, backward chaining has a smaller search space and thus is more efficient~\cite{LAMBADA22}. Moreover,~\cite{LAMBADA22} proposed a backward chaining modular framework with LLMs as modules, which attains better performance than the existing forward chaining frameworks. Another direction is to perform forward chaining (deduction) and backward chaining (abduction) simultaneously~\cite{METGEN22}. In addition to proof-finding, backward chaining can also be generalized to more general problems. For example,~\cite{Entailer22} applied it to a multi-choice question-answering problem by combing the question and each answer choice into a verifiable hypothesis.

However, researches on this line are more under-explored than forward chaining.

\subsubsection{Question Decomposition} 
Question decomposition is a backward reasoning method to improve performance on multi-hop questions that require integrating multiple pieces of knowledge and inferring over them to obtain the answers. It decomposes each question into several simpler sub-questions and answers these sub-questions to derive the final answers. In analogy to forward reasoning, solving a single-hop sub-question is to query a single piece of knowledge, and combining sub-answers to form the final answer is inference. And decomposing a question into sub-questions is an abductive step. In other words, while question decomposition introduces abduction steps, it removes the requirement of multi-step knowledge selection/retrieval. 

Multi-hop questions are difficult to answer because they have a long tail distribution and are challenging to find the relevant multiple pieces of knowledge. Especially, it might be very challenging to find the required knowledge for implicit multi-hop questions, whose superficial text and semantics can be very different to the required knowledge. By contrast, it is easier to query a piece of knowledge and answer each decomposed single-hop sub-question. For example,~\cite{QD22} demonstrated that both medium-size PLMs and LLMs can significantly improve the performance on multi-hop questions with human-decomposed questions. It was also found effective in mathematical reasoning and symbolic reasoning~\cite{Least-to-Most22}. Besides, previous research has also shown that question decomposition is effective with both medium-size PLMs~\cite{LDHQD22} and LLMs~\cite{self-ask22} on multi-hop questions. Research of this line has a longer history than LLM-only CoT methods.

\paragraph*{Decomposition of explicit and implicit multi-hop question} 
According to the difficulty of decomposition, multi-hop questions can be divided into explicit multi-hop questions and implicit multi-hop questions. Explicit multi-hop questions are those which can be decomposed simply based on their superficial text (syntactical pattern). For example, the question ``where was Obama's wife born?'' can be decomposed into ``who is Obama's wife?'' and ``where was \#1 born?''\footnote{``\#1'' denotes the answer of the first sub-question.} based on the superficial text of the original question. Implicit multi-hop questions, however, are more difficult to decompose since their sub-questions are not syntactically consistent with the questions. For example, the question ``can we directly live in the space?'' needs to be decomposed into ``what do we need to keep alive?'' and ``are there \#1 in the space?'', where the key predicate in the first sub-question ``need'' is not explicitly mentioned in the original question. While explicit multi-hop questions can be decomposed based on their superficial text and syntactical structures via extraction and editing~\cite{DecompRC19}, decomposing implicit multi-hop questions is much more difficult. A key challenge is that it lacks large-scale annotated data, which is labour-intensive to obtain especially as the number of hops increases. StrategyQA~\cite{StrategyQA21} is an implicit multi-hop question dataset annotated with sub-questions and the corresponding knowledge pieces, but its size is small (2.7k). To alleviate the data sparsity problem, there is some research on weak supervision data~\cite{ONUS20,LDHQD22}. Recently, in-context learning provides a new solution~\cite{QD22,self-ask22} to this problem, which requires only a small set of demonstrations.

\paragraph*{Framework with respect to sequential and tree structure} 
There are different structures of decomposition based on the dependencies among the parent question and sub-questions, involving sequential structure and tree structure. In a sequential structure, each sub-question is linearly dependent on the answer (e.g. a bridge entity) of the antecedent sub-question, and the answer of the last sub-question is the multi-hop question's answer. For example, the answer ``Michelle'' of the first sub-question ``who is Obama's wife?'' makes up of its subsequent sub-question ``where was \#1 born?'' whose answer ``1964'' is also the final answer of the multi-hop question ``where was Obama's wife born?''. By contrast, in a tree structure, sub-questions are independent to each other with their answers equally contributing to the final answer. For example, the question ``who can swim better, elephant or dolphin?'' consists of ``can elephant swim?'' and ``can dolphin swim?'', and the final answer is derived by composing the corresponding sub-answer ``elephant can't swim'' and ``dolphin can swim''. There are three kinds of decomposition-based framework: module-based decomposition~\cite{DecompRC19}, decompose-then-recompose~\cite{ONUS20,QD22}, and generate-then-answer~\cite{modularqa21,self-ask22}. The first framework designs different modules responsible for different reasoning types, which separate and model sequential and tree structure independently~\cite{DecompRC19}. The decompose-then-recompose framework first decomposes the multi-hop question into all its comprised sub-questions and recomposes their sub-answers to derive the final answer~\cite{ONUS20,QD22}. However, it ignores the dependencies among sub-questions (sequential structure). By contrast, the last one, generate-then-answer, is sequential in nature, which iteratively generates and answers a single-hop sub-question~\cite{modularqa21,self-ask22}. It considers the question dependencies in sequential structure and is compatible with tree structure, but is less efficient than decompose-then-recompose since it can't solve sub-questions of tree structure in parallel. 

However, it is still challenging to solve multi-hop questions with very long hops. Due to the combinatorial explosion, it becomes increasingly difficult to annotate decomposition supervision data and provide representative demonstrations for in-context learning. Also, there are likely to exist multiple decomposition paths when there are long hops, which also puts a challenge on planning. The following are the potential directions we suggest.
\begin{itemize}
    \item \textbf{Hierarchical decomposition}. Instead of directly decomposing the multi-hop question into the simplest single-hop sub-questions, it might be easier for models to perform hierarchical decomposition, i.e. repeatedly decompose multi-hop questions into simpler multi-hop questions until there are only single-hop questions. Moreover, it is also more practical for researchers to annotate supervision data or select appropriate exemplars of in-context learning for layer-by-layer decomposition. 
    \item \textbf{Knowledge-aware planning}. When there exist multiple decomposition paths, it is critical to plan for a decomposition way to answerable sub-questions. For this purpose, it is important to be aware of what the existing knowledge there is.
\end{itemize}

\subsection{Summary}
Reasoning requires models to integrate multiple knowledge and reason over them. Early research mostly improved reasoning performance via architectural designs and only constructed forward reasoning paths for interpretability or faithfulness. Specialized models were designed to improve evidence aggregation, reasoning capability or faithfulness, but they are constrained to specific tasks, datasets or reasoning types that hurt the generalization. Since transformers have been found to be soft deductive reasoners after in-domain finetuning, vanilla PLMs have been more popular to perform reasoning. However, data sparsity and spurious correlation problems make it difficult for medium-size PLMs to learn the general logical structure of diverse reasoning types. There are also some researches incorporating inductive biases via specialized pretraining, but it is unclear whether this is still worth as the model size and the number of pretraining data increases. Recently, it was found that an emergent ability comes as PLMs are large enough: generating a reasoning path before the final answer can significantly improve the multi-step reasoning performance, which boosts much research on this line. In addition to the forward reasoning direction, the other reasoning direction is backward, which is more efficient than forward reasoning due to the smaller search space. While forward reasoning can expose arbitrary new knowledge entailed by the existing knowledge, backward reasoning just targets at the specific goal or the problem solution. A typical approach of backward reasoning is question decomposition, which can improve performance on multi-hop questions for both medium-size PLMs and LLMs. While there is much research on deductive reasoning, defeasible reasoning is much more challenging for PLMs and is still under-explored.

\section{NLR Benchmarks}\label{sec:topics}
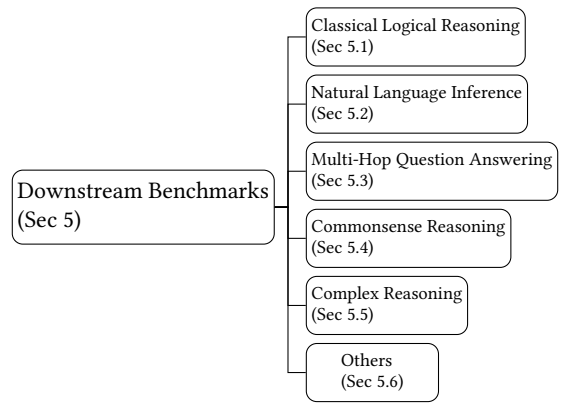
\begin{figure*}[tp]
    \centering
    \begin{forest}
        forked edges,  
        for tree={
            grow=east,
            anchor=base west,  
            reversed=true,
            font=\small,
            align=left,
            draw=black,  
            rounded corners,  
            minimum width=5em,
            s sep=3pt,
            inner xsep=2pt,  
            inner ysep=3pt,  
        },
        where level=1{font=\scriptsize,}{},
        where level=2{font=\scriptsize,}{},
        where level=3{font=\scriptsize,}{},
        [
            Downstream Benchmarks \\ (Sec~\ref{sec:topics})
            [Classical Logical Reasoning \\ (Sec~\ref{sec:CLR})]
            [Natural Language Inference \\ (Sec~\ref{sec:NLI})]
            [Multi-Hop Question Answering \\ (Sec~\ref{sec:MHQA})]
            [Commonsense Reasoning \\ (Sec~\ref{sec:CR})]
            [Complex Reasoning \\ (Sec~\ref{sec:complex})]
            [Others \\ (Sec~\ref{sec:others})]
        ]
    \end{forest}
    \caption{Natural language reasoning benchmarks in NLP}
    \label{fig:cate_nlr}
\end{figure*}

In this section, we review some typical and popular downstream benchmarks thought to require natural language reasoning and discuss to what extent they are actually related to reasoning. Although there might be more downstream benchmarks with respect to natural language reasoning, here we mainly focus on four of the most popular and familiar to the community: classical logical reasoning, natural language inference, multi-hop question answering, and commonsense reasoning. We list the corresponding datasets and benchmarks and briefly introduce the development. Besides, we present some datasets collected from realistic examinations or explicitly designed to challenge LLMs, which we name ``complex reasoning''. In addition to well-known reasoning benchmarks, we also introduce some other tasks that require performing natural language reasoning. A figure of the taxonomy is shown in Fig~\ref{fig:cate_nlr}.

\subsection{Classical Logical Reasoning}\label{sec:CLR}
Some datasets explicitly target classical reasoning types in philosophy and logic, e.g. deduction, abduction and induction, following the definitions in the two areas. Thus, we call them ``classical logical reasoning tasks''. A key characteristic of this topic is that tasks are mostly artificial to study reasoning. There are both deductive reasoning and defeasible reasoning.

\subsubsection{Deductive reasoning} 
Classical deductive reasoning tasks are defined formally based on formal logic, such as propositional logic and first-order logic. There are mainly three types of task: inference~\cite{bAbI16,AAC21,LGID22}, theorem proving~\cite{ruletaker20,ProofWriter21,PARARULE-Plus22,FOLIO22} and reasoning path generation~\cite{LGID22}. The inference task is to reason the conclusion given the premises in a single step, while theorem proving is to predict whether the given proposition is true or false with the given knowledge bases, which usually requires multiple steps. Obviously, inference is the fundamental task that forms the basic capability of multi-step reasoning tasks such as theorem proving, while reasoning path generation is an interpretable task that can be complementary to multi-step reasoning. However, except FOLIO~\cite{FOLIO22}, all the existing explicit deductive reasoning datasets are synthesized. We list the classical deductive reasoning datasets in Table~\ref{tab:dedu_data}.

\begin{table}[!ht]
    \centering\scriptsize
    \begin{tabular}{lllll}
    \toprule
        \textbf{Dataset}         & \textbf{Size}   & \textbf{Data Source}      & \textbf{Task}   & \textbf{Remark}               \\ 
        \midrule
        \textbf{bAbI-15}~\cite{bAbI16}     & -        & synthetic             & inference   & basic deduction     \\ 
        \hline
        \textbf{RuleTaker}$\dagger$~\cite{ruletaker20}/\textbf{ProofWriter}$\dagger$~\cite{ProofWriter21}        & 500k     & synthetic        & theorem proving   & the first natural language theorem proving    \\ 
        \hline
        \textbf{PARARULE-Plus}~\cite{PARARULE-Plus22}        & 400k     & synthetic        & theorem proving   & addresses the depth imbalance issue on ParaRules    \\ 
        \hline
        \textbf{AAC}~\cite{AAC21}              & 710k             & synthetic         & inference   & based on 8 syllogistic argument schemes          \\ 
        \hline
        \textbf{LogicInference}~\cite{LGID22}       & 200k        & synthetic         & \makecell[l]{inference\\reasoning path generation}  & -            \\ 
        \hline
        \textbf{FOLIO}~\cite{FOLIO22}         & 1.4k         & expert-written         & theorem proving   & more diverse patterns    \\ 
        \bottomrule
    \end{tabular}
    \caption{\label{tab:dedu_data}Datasets of classical deductive reasoning, where bAbI-15 means ``the 15-th task in bAbI tasks''. $\dagger$ denotes there are ground reasoning paths.}
\end{table}

\paragraph*{Proof-finding and faithful reasoning} 
Since ~\cite{ruletaker20} has proposed a theorem proving dataset and showed that vanilla medium-size PLMs can be soft theorem provers, a series of researches emerge on this task to study natural language reasoning, with both vanilla medium-size PLMs~\cite{PRover20,multiPRover21,FaiRR22,NLProofS22,paradox22} and LLMs~\cite{ProofWriter21,selection-inference22,faithful22,PRONTOQA22,LAMBADA22}. However, while the performance of transformers on theorem proving is promising, ~\cite{paradox22} found that there are some statistical features inherently existing in the problem, which may hinder models from generalization. In addition to just classifying the final label\cite{ruletaker20,paradox22}, it has been demonstrated that producing proofs can bring better generalization ability to unseen proof depth and out-of-domain data~\cite{PRover20,ProofWriter21} and contribute to interpretability. There is several research on proof generation or proof-finding, either forward~\cite{ProofWriter21,FaiRR22,selection-inference22,faithful22} or backward~\cite{EVR21,IBR22,LAMBADA22}, where backward chaining is more efficient than forward chaining on proof-finding intrinsically~\cite{LAMBADA22}. To alleviate the combinatorial explosion problem in the search space of the forward chaining, some researchers proposed planning~\cite{faithful22,NLProofS22}. Moreover, faithful reasoning is also an interesting topic in this problem, where the procedure of reasoning is strictly designed to guarantee that models actually perform reasoning to derive the answer rather than rely on shortcuts~\cite{FaiRR22,faithful22}. However, while the performance is promising, even approaching perfect sometimes, all research mentioned above is based on synthetic datasets. Moreover, recently, the new expert-written dataset FOLIO~\cite{FOLIO22} showed that when it comes to more diverse natural language, the performance degrades severely. By contrast, the entailment tree generation dataset EntailmentBank~\cite{EntailmentBank21} is often used to study the proof generation and faithful reasoning as with theorem proving~\cite{NLProofS22,faithful22,METGEN22,IRGR22,Entailer22}. The target hypotheses in this dataset are collected from realistic examinations and proofs are annotated by humans, which is a better alternative for studies on proof generation.

There are also some benchmarks to diagnose model's capabilities on logical semantics understanding~\cite{ConjNLI20,RobustLR22,CONDAQA22}.

\subsubsection{Defeasible reasoning} 
Two typical defeasible reasoning types are abduction~\cite{ART20,AbductionRules22} and induction~\cite{bAbI16,DEER22}. There is also another type of defeasible reasoning~\cite{defeasibleNLI20}. Datasets are shown in Table~\ref{tab:defea_data}. Compared to classical deductive reasoning, researches on defeasible reasoning are still under-explored. Experiments suggested that there remains a large space to improve~\cite{DEER22}.

\paragraph*{Inductive reasoning} 
Induction produces a more general principle from the given knowledge that can express or explains them. Early datasets require first inducing rules and then applying them to perform deduction, without inducing explicit rules~\cite{bAbI16,CLUTRR19}. Recently, a new dataset DEER~\cite{DEER22} studies rule prediction, where the task is to induce natural language rules from natural language facts.

\paragraph*{Abductive reasoning} 
Abduction is to predict the best explanation for the observations. According to the mode of the reversed reasoning, abduction can provide explanations that constitute premises of whether deductive reasoning~\cite{AbductionRules22} or defeasible reasoning~\cite{ART20}. Based on the explained objects (i.e. input), abduction may target a small set of premises~\cite{ART20} or a knowledge corpus~\cite{AbductionRules22}. 

\paragraph*{Others} 
In addition to abduction and induction, defeasibleNLI~\cite{defeasibleNLI20} focuses on whether a premise can weaken or strengthen a probable conclusion. There are researches on defeasible inference graphs to improve both human reasoning~\cite{defeasiblegraphs21} and machine reasoning performance~\cite{thinkaboutit21}.

\begin{table}[!ht]
    \centering\footnotesize
    \begin{tabular}{llllll}
    \toprule
        \textbf{Dataset}               & \textbf{Reasoning}            & \textbf{Size}             & \textbf{Source}              & \textbf{Task}                   & \textbf{Remark}                  \\ 
        \midrule
        \textbf{bAbI-16}~\cite{bAbI16}               & induction            & -                & synthetic                   & extraction                & induce-then-deduce             \\ 
        \hline
        \textbf{CLUTRR}~\cite{CLUTRR19}                  & induction            & -             & synthetic                   & extractive QA                & induce-then-deduce       \\ 
        \hline
        \textbf{DEER}~\cite{DEER22}                  & induction            & 1.2k             & Wikipedia                   & generation                & rule prediction       \\ 
        \hline
        \textbf{AbductionRules}~\cite{AbductionRules22}        & abduction            & -                & synthetic                   & generation                & abduce from knowledge database \\ 
        \hline
        \textbf{ART}~\cite{ART20}                   & abduction            & 17.8k            & ROCStories~\cite{ROCStories16}                  & 2-choice/generation       & abduce from two premises \\
        \hline
        \textbf{defeasibleNLI}~\cite{defeasibleNLI20}         & others               & 43.8k            & other datasets              & classification/generation & concern the change of strength   \\ 
        \bottomrule
    \end{tabular}
    \caption{\label{tab:defea_data}Datasets of classical defeasible reasoning, where bAbI-16 means ``the 16-th task in bAbI tasks''.}
\end{table}

\subsection{Natural Language Inference}\label{sec:NLI}
Natural language inference (NLI), also known as recognizing textual entailment (RTE), is a typical task in NLP. It is a 3-way classification task labelling as entailment, contradiction and neutral, to identify whether the given premise entails a hypothesis. An entailment is described as a conclusion that a person would typically infer from the premise or the implication described by the premise~\cite{RTE13,SNLI15}. 

While NLI is regarded as a natural language understanding~\cite{SNLI15} or natural language reasoning~\cite{ruletaker20} problem, we find it involves examples of both understanding and reasoning problems. Specifically, we identify there are mainly three types of premise-hypothesis entailment problems: paraphrasing, compound semantics understanding, and reasoning with implicit premises. For the first type, the hypothesis is a paraphrase of the premise. For the second type, the premise is a compound proposition entailing the hypothesis. For the last type, there need some unstated premises to link the provided premise to the hypothesis. We demonstrate samples from the popular dataset SNLI~\cite{SNLI15} for each type respectively in Table~\ref{tab:type_of_entail}.

\begin{table}[!ht]
    \begin{tabular}{lll}
    \toprule
        \textbf{}           & \textbf{Premise}                                      & \textbf{Hypothesis}             \\ 
        \midrule
        \textbf{Paraphrase} & Two doctors perform surgery on patient               & Doctors are performing surgery \\ 
        \hline
        \textbf{CSU}        & Two women are embracing while holding to go packages & \makecell[l]{Two women are holding packages\\\orange{(Two women are embracing)}} \\ 
        \hline
        \textbf{Reasoning}  & \makecell[l]{A soccer game with multiple males playing\\\blue{(Soccer is a sport)}}            & Some men are playing a sport   \\
    \bottomrule
    \end{tabular}
    \caption{\label{tab:type_of_entail}Examples from SNLI~\cite{SNLI15} of three types of entailment, where CSU indicates ``Compound Semantics Understanding''. The blue-coloured sentence is the implicit premise, while the orange-coloured sentence is the other semantics of the premise.}
\end{table}

There are several popular generic datasets listed in Table~\ref{tab:NLI_reason_data}, where datasets with realistic hypotheses have few hypothesis-only biases than those with human-authored hypotheses. Concretely, it has been found that there are significant biases in human-authored hypotheses~\cite{SNLI15,MultiNLI18}, with which models can even predict the label without premise~\cite{hypo-bias-SNLI18,Human-Elicited-hypo-bias18,gen-debiased-nli22}.

\begin{table}[!ht]
    \centering \footnotesize
    \begin{tabular}{llllll}
    \toprule
        \textbf{Dataset}     & \textbf{Domain}  & \textbf{Size} & \textbf{P Source}                & \textbf{H Source} & \textbf{Remark}               \\ 
        \midrule
        \textbf{SNLI}~\cite{SNLI15}/\textbf{e-SNLI}$\dagger$~\cite{eSNLI18}        & generic & 570k & realistic                        & human-authored & the first large-scale NLI dataset                  \\ 
        \hline
        \textbf{MultiNLI}~\cite{MultiNLI18}    & generic & 433k & realistic                    & human-authored           & cover more styles and topics        \\ 
        \hline
        \textbf{XNLI}~\cite{XNLI18}    & generic & 7.5k & -                    & -           & cross-lingual, based on MultiNLI        \\ 
        \hline
        \textbf{SciTail}~\cite{SciTail18}     & science & 27k  & realistic                  & realistic                      & the first NLI dataset with entirely realistic data         \\ 
        \hline
        \textbf{SciNLI}~\cite{SciNLI22}      & science & 107k & realistic              & realistic          & -       \\ 
    \bottomrule
    \end{tabular}
    \caption{\label{tab:NLI_reason_data}Datasets of NLI. ``P'' denotes ``Premise'' while ``H'' denotes ``Hypothesis''. $\dagger$ means that e-SNLI provides explanations for examples of SNLI.}
\end{table}

Several datasets and benchmarks of NLI are just understanding problems, such as those presented specifically to probe and improve the model capabilities of paraphrase and compound semantics understanding~\cite{HELP19,MED19,ConjNLI20,mosharafhossain20}. Also, datasets that are converted from other tasks into NLI-style are irrelevant to reasoning when they are not the reasoning problems originally~\cite{Figurative-NLI21,DocNLI21}.

Interestingly, it was shown that crowdworkers sometimes annotated different labels to the same premise-hypothesis pair~\cite{SNLI15,UNLI20}. We think this phenomenon can be attributed to the existence of defeasible reasoning, where people with different background knowledge can derive different conclusions.


\subsection{Multi-Hop Question Answering}\label{sec:MHQA}
Multi-hop question answering (MHQA) studies answering the complex questions that require reasoning over evidence scattered in different contexts\footnote{There are not only natural language reasoning questions, but also other types such as numerical comparison~\cite{HotpotQA18,StrategyQA21}.}, thus it is also called as multi-hop reading comprehension, where candidate contexts are either explicitly provided involving some distractors~\cite{WikiHop-MedHop18,HotpotQA18,2wikimultihop20,musique22} (distractor setting), or can be retrieved from external knowledge bases such as Wikipedia~\cite{HotpotQA18,BeerQA21,StrategyQA21} and WorldTree~\cite{QASC20} (retrieval setting). The term ``hop'' here indicates the number of contexts required to reason rather than the number of inference steps, which describes the behaviour moving among different contexts.

\paragraph*{Datasets \& Benchmarks} 
We list some typical datasets in Table~\ref{tab:MHQA_data}. A key challenge on dataset construction is that it is very label-intensive to annotate large-scale multi-hop questions especially there is a combinatorial explosion as the number of hops increases. Many datasets are synthetic or semi-synthetic~\cite{WikiHop-MedHop18,HotpotQA18,R4C20,2wikimultihop20}, where questions are mainly deductive, i.e. the answers are necessarily true with the given contexts. There are two types of rationale: supporting text set~\cite{HotpotQA18,2wikimultihop20,musique22,QASC20,StrategyQA21} and reasoning path including both forward~\cite{R4C20,eQASC20} and backward~\cite{musique22,StrategyQA21}.

\begin{table}[!ht]
    \centering \small
    \begin{tabular}{lllllll}
    \toprule
        \textbf{Dataset}         & \textbf{Domain}   & \textbf{Size} & \textbf{CS} & \textbf{QS} & \textbf{AT}  & \textbf{Rationale}     \\ \hline
        \textbf{WikiHop}~\cite{WikiHop-MedHop18}         & generic           & 51k            & \blue{Wikipedia} & synthetic & option & $\times$                          \\ 
        \midrule
        \textbf{MedHop}~\cite{WikiHop-MedHop18}          & medicine & 2.5k                & \blue{Medline} & synthetic & option   & $\times$                          \\ 
        \hline
        \textbf{HotpotQA}~\cite{HotpotQA18}        & generic           & 112k             & Wikipedia & semi-synthetic & \makecell[l]{span\\yes/no} & sentences                  \\ 
        \hline
        \textbf{R4C}~\cite{R4C20}             & generic           & 4.6k            & Wikipedia & semi-synthetic & \makecell[l]{span\\yes/no} & triples           \\ 
        \hline
        \textbf{BeerQA}~\cite{BeerQA21}          & generic           & 530             & \orange{Wikipedia} & human-authored & \makecell[l]{span\\yes/no} & $\times$                          \\ 
        \hline
        \textbf{2WikiMultiHopQA}~\cite{2wikimultihop20} & generic           & 192k              & \blue{Wikipedia}  & synthetic & span & \makecell[l]{sentences\\triples}          \\ 
        \hline
        \textbf{MuSiQue}~\cite{musique22}         & generic           & 25k                  & \blue{Wikipedia} & human-composed & span & \makecell[l]{paragraphs\\decomposition$\star$}                 \\ 
        \hline
        \textbf{QASC}~\cite{QASC20}/\textbf{eQASC}$\dagger$~\cite{eQASC20}            & science           & 9.9k          & \orange{WorldTree} & human-authored & option & \makecell[l]{sentences\\ reasoning path~\cite{eQASC20}$\star$} \\ 
        \hline
        \textbf{StrategyQA}~\cite{StrategyQA21}      & generic           & 2.7k          & \orange{Wikipedia} & human-authored  & yes/no  & \makecell[l]{paragraphs\\decomposition$\star$}                 \\
        \bottomrule
    \end{tabular}
    \caption{\label{tab:MHQA_data}Datasets of multi-hop question answering. $\dagger$ indicates it annotates the rationale for this dataset. ``CS'' denotes ``Context Source'', ``QS'' denotes ``Question Source'', and ``AT'' denotes ``Answer Type''. In CS, the distractor setting is coloured blue, while the retrieval setting is coloured orange, and black means there are both. For rationale, $\star$ means ``reasoning path'', otherwise ``supporting evidence set''. ``decomposition'' indicates the ground annotations of decomposed sub-questions and the corresponding contexts.}
\end{table}

\paragraph*{Multi-hop question construction} 
There are mainly two lines on multi-hop question construction: improve data quality and increase data number. Firstly, it has been found that there are artifacts in HotpotQA that can be leveraged to answer questions without performing multi-hop reasoning~\cite{Adversarial-MultiHopQA19,dataset-design-choice19,single-hop-rc19,DiRe20}. To deal with this problem, one way is to leverage adversarial data~\cite{2wikimultihop20}, another way is to construct new datasets of high-quality multi-hop questions with carefully designed data collection strategies~\cite{2wikimultihop20,musique22}. Secondly, as multi-hop questions are difficult to annotate, there are some researches on automatic data generation\cite{MHGM20,unsup21,CQG22}.

\paragraph*{Reasoning} 
After deriving the relevant contexts, it requires aggregating multiple pieces of evidence and reasoning over them. Firstly, there are some specialized models designed for better evidence aggregation~\cite{BFR-Graph21,Transformer-XH20}. Secondly, reasoning is usually performed via end-to-end answering~\cite{SRLGRN20,Transformer-XH20,BFR-Graph21,pseudo-envidence21,suqa21} or backward decomposition~\cite{DecompRC19,ONUS20,StrategyQA21,modularqa21,QD22,self-ask22}. In this topic, question decomposition (i.e. backward reasoning) is more popular than forward reasoning.

\subsection{Commonsense Reasoning}\label{sec:CR}
Commonsense reasoning deals with implicit commonsense knowledge, where commonsense knowledge is necessarily required to solve the problem. Such knowledge may be obvious to people but non-trivial to machines since they are difficult to retrieve from the web due to reporting bias, e.g. ``when people are hungry, they would like to eat something''. 

However, although it is named as ``commonsense reasoning'', not all the datasets are reasoning as defined (Sec~\ref{sec:def}), such as querying shared living experiences~\cite{ProtoQA20}, identifying pragmatic implications~\cite{SBIC20}, and so on~\cite{CommonGen20,MRF22}.

\subsubsection{Datasets \& Benchmarks}

According to the conclusion type, there are mainly three types of reasoning problems in commonsense reasoning: ``what'' (i.e. assertions or events) ``what if / why'' (e.g. causal and temporal relations between events), and ``how'' (i.e. actions). 

\paragraph*{What} 
This type of problem is similar to multi-hop question answering, where the problems require combining multiple pieces of knowledge that some are from external knowledge sources. The key difference is that it requires some commonsense knowledge, which is not explicitly provided, in commonsense reasoning. In other words, the problems require integrating explicit knowledge, such as science~\cite{OpenBookQA18,OpenCSR21}, with some commonsense knowledge. We list some datasets in Table~\ref{tab:CR_assertion_data}. 

\begin{table}[!ht]
    \centering\small
    \begin{tabular}{llllllll}
    \toprule
        \textbf{Dataset}    & \textbf{Other Knowledge} & \textbf{Knowledge Source}        & \textbf{Size} & \textbf{Task}      & \textbf{Rationale}           \\ 
        \midrule
        \textbf{OpenBookQA}~\cite{OpenBookQA18} & science                  & WorldTree                        & 6k            & multi-choice QA    & science facts            \\ 
        \hline
        \textbf{OpenCSR}~\cite{OpenCSR21}    & science                  & WorldTree, ARC corpus            & 20k           & free-form QA       & $\times$   \\ 
        \hline
        \textbf{CREAK}~\cite{CREAK21}      & entity                   & Wikipedia                        & 13k           & claim verification & explanation  \\ 
        \bottomrule
    \end{tabular}
    \caption{\label{tab:CR_assertion_data}Datasets of ``what'' commonsense reasoning.}
\end{table}

\paragraph*{What if \& Why} 
This type of problem often reasons for causal and temporal relations between events. There are two causal relations: causes and effects, which can be seen as backward causal reasoning and forward causal reasoning respectively. Take the causality of events as an example, forward causal reasoning asks ``what events are likely to happen next?'', while backward causal reasoning asks ``what may cause this event?'' in a scenario described by the context, i.e. querying the plausible previous or subsequent events respectively. Besides, there are some problems that require considering another scenario in addition to the context, which can be seen as constrained causal reasoning. For example, TIMETRAVEL~\cite{TIMETRAVEL19,ART20} is a counterfactual story rewriting dataset, where the original story is also given. See relevant datasets and benchmarks in Table~\ref{tab:CR_event_data}.

\begin{table}[!ht]
    \centering\footnotesize
    \begin{tabular}{llllll}
    \toprule
        \textbf{Dataset}    & \textbf{Size} & \textbf{Direction}  & \textbf{Context Source} & \textbf{Task}  & \textbf{Remark}  \\ 
        \midrule
        \textbf{ROCStories}~\cite{ROCStories16}       & 50k            & temporal        & human-authored          & 2-choice QA  & -                   \\ 
        \hline
        \textbf{SWAG}~\cite{SWAG18}       & 113k            & temporal        & ActivityNet, LSMDC          & multi-choice QA  & -                   \\ 
        \hline
        \textbf{HellaSwag}~\cite{HellaSwag19}       & 20k            & temporal        & ActivityNet, WikiHow          & multi-choice QA  & an upgraded SWAG                   \\ 
        \hline
        \textbf{COPA}~\cite{COPA11}       & 1k            & both        & human-authored          & 2-choice QA  & -                   \\ 
        \hline
        \textbf{Social-IQA}~\cite{Social-IQA19} & 38k           & both       & human-authored          & multi-choice QA  & social situations                  \\ 
        \hline
        \textbf{e-CARE}$\dagger$~\cite{e-CARE22}     & 21k           & both       & human-authored          & 2-choice QA  & -   \\ 
        \hline
        \textbf{WIQA}~\cite{WIQA19}     & 40k           & forward       & ProPara~\cite{ProPara18}     & multi-choice QA  & about nature processes    \\ 
        \hline
        \textbf{TIMETRAVEL}~\cite{TIMETRAVEL19} & 29k           & forward       & ROCStories~\cite{ROCStories16}              & generation   & counterfactual reasoning                     \\ 
        \hline
        \textbf{ART}~\cite{ART20}             & 20k           & backward      & ROCStories~\cite{ROCStories16}             & 2-choice/generation  & abductive commonsense reasoning \\
        \hline
        \textbf{TellMeWhy}~\cite{tellmewhy21}  & 30k           & backward            & ROCStories~\cite{ROCStories16}              & free-form QA    & each annotated 3 possible answers                  \\ 
        \hline
        \textbf{WikiWhy}$\dagger$~\cite{WikiWhy22}      & 9k   & backward      & human-edited Wikipedia  & free-form QA  & about Wikipedia entities / events   \\
    \bottomrule
    \end{tabular}
    \caption{\label{tab:CR_event_data}Datasets of ``what if'' / ``why'' commonsense reasoning, where $\dagger$ denotes there annotates supporting facts or reasoning paths. For direction, ``both'' indicates there are both forward and backward causal reasoning.}
\end{table}

\paragraph*{How}
This type of problem is mainly about ``how to do it''. It is more complex and also involves problem-solving and decision-making. See some in Table~\ref{tab:CR_action_data}).

\begin{table}[!ht]
    \centering\footnotesize
    \begin{tabular}{llllll}
    \toprule
        \textbf{Dataset}    & \textbf{Size} & \textbf{Context Source} & \textbf{Option Source} & \textbf{Task}  & \textbf{Remark}  \\ 
        \midrule
        \textbf{WikiHow Goal-Step}~\cite{wikihow-goal-step20}       & 1489k    & WikiHow       & automatically generated          & multi-choice  & goals, steps, and temporal ordering                   \\ 
        \hline
        \textbf{PIQA}~\cite{PIQA20} & 21k      & human-authored     & human-authored      & 2-choice  & physical causal reasoning                  \\ 
    \bottomrule
    \end{tabular}
    \caption{\label{tab:CR_action_data}Datasets of ``how'' commonsense reasoning.}
\end{table}

\paragraph*{Others}

Besides, some datasets involve multiple types of reasoning. We list some typical datasets in Table~\ref{tab:CR_hybrid_data}. 

\begin{table}[ht]
    \centering\scriptsize
    \begin{tabular}{llllll}
    \toprule
                    & Size & Context Source & Question Source & Task & Remark \\
        \midrule
        \makecell[l]{\textbf{CSQA}~\cite{CQA19}\\\textbf{CoS-E}$\dagger$~\cite{CoS-E19}/\textbf{ECQA}$\dagger$~\cite{ECQA21}} & 12k & - & semi-synthetic   & multi-choice QA   &  \makecell[l]{ConceptNet concepts~\cite{ConceptNet17}\\explanation~\cite{CoS-E19,ECQA21}, commonsense facts~\cite{ECQA21}}   \\
        \hline
        \textbf{CSQA2}~\cite{CSQA2-21} & 14k   & - & human-authored   & boolen QA  & data construction via gamification \\
        \hline
        \textbf{CosmosQA}~\cite{CosmosQA19}        & 35k            &  blog~\cite{Spinn3r09}  &  human-authored        & multi-choice QA & reading comprehension on blogs  \\
        \hline
        \textbf{Moral Stories}~\cite{moral-stories21} & 12k & human-authored & - & classification/generation & situated reasoning with social norms \\
    \bottomrule
    \end{tabular}
    \caption{\label{tab:CR_hybrid_data}Datasets and benchmarks with multiple types of commonsense reasoning. $\dagger$ indicates it annotates the rationale for the dataset.}
\end{table}

\subsubsection{Reasoning}
Since commonsense knowledge is essential to this topic, much research focused on commonsense knowledge~\cite{CYC95,ConceptNet04,ConceptNet17,Event2Mind18,ATOMIC19,ATOMIC20,social-norm20,commonsense-bert19}. As for the reasoning system, there are mainly two types of methods: graph-based~\cite{MHGRN20,QA-GNN21,OpenCSR21,GreaseLM22} and vanilla PLMs~\cite{commonsense-bert19,PLM-CR20,DELOREAN20,ALERT22,GKP22,MAIEUTIC22}, where graph-based methods are designed to aggregate knowledge from commonsense knowledge bases while vanilla PLMs are used as implicit knowledge bases themselves.

\subsection{Complex Reasoning}\label{sec:complex}
There are some datasets collected from realistic examinations or tests, which may require domain-specific knowledge and multiple types of reasoning skills (Table~\ref{tab:realistic_data}). 

\begin{table}[!ht]
    \centering\footnotesize
    \begin{tabular}{llllll}
    \toprule
        \textbf{Dataset} & \textbf{Size} & \textbf{Domain} & \textbf{Source}                              & \textbf{Task}        \\ 
        \midrule
        \textbf{AR-LSAT}~\cite{AR-LSAT22} & 2k            & law             & law school admission test                    & multi-choice QA      \\ 
        \hline
        \textbf{HEAD-QA}~\cite{HEAD-QA19} & 6.7k          & healthcare      & specialized healthcare examination           & multi-choice QA      \\ 
        \hline
        \textbf{AI2-ARC}~\cite{AI2-ARC18}/\textbf{EntailmentBank}$\dagger$~\cite{EntailmentBank21} & 7.7k          & science         & grade-school standardized test               & multi-choice QA      \\ 
        \hline
        \textbf{ReClor}~\cite{ReClor20}/\textbf{MetaLogic}$\dagger$~\cite{MetaLogic22}  & 6k            & generic         & standardized graduate admission examination  & RC + multi-choice QA \\ 
        \hline
        \textbf{LogiQA}~\cite{LogiQA20}  & 8k            & generic         & national civil servants examination of China & RC + multi-choice QA \\ 
        \hline
        \textbf{ConTRoL}~\cite{ConTRoL21} & 8k            & generic         & competitive selection and recruitment test   & passage-level NLI    \\ 
    \bottomrule
    \end{tabular}
    \caption{\label{tab:realistic_data}Complex reasoning datasets with the realistic data from examinations or tests, where ``RC'' denotes ``reading comprehension''. $\dagger$ indicates ``it annotates reasoning paths for some examples in this dataset''.}
\end{table}

To better diagnose the ability of LLMs, two few-shot prompting benchmarks called MMLU~\cite{MMLU21} and Big-Bench~\cite{BIG-bench22} are proposed, where tasks are much more challenging and even believed to be beyond the capabilities of current language models, in which some require to perform reasoning. Among tasks in Big-Bench, ~\cite{BBH22} identified 23 challenging tasks, named as Big-Bench Hard (BBH), that LLMs failed to surpass the average human-rater, and many of them require to perform multi-step reasoning. However, when equipped with CoT prompting, GPT3 can outperform human performance on a major of these hard tasks.

\subsection{Others}\label{sec:others}
In addition to the above-mentioned datasets and benchmarks, there are also some other tasks requiring natural language reasoning scattered in the NLP domain, involving dialog~\cite{ShARC18}, reading comprehension~\cite{ROPES19} and so on~\cite{ARC18}. Note that reasoning is an important method to arrive at the required answers or solutions, which is of more frequent usage in complex problems. In other words, reasoning can occur in many other domains to solve challenging problems that require multiple knowledge to derive conclusions. While there might be more reasoning tasks or datasets, we just list some of them in Table~\ref{tab:other_data}.

\begin{table}[!ht]
    \centering
    \begin{tabular}{llllll}
    \toprule
        \textbf{Dataset}        & \textbf{Size} & \textbf{Reasoning}        & \textbf{Context Source}     & \textbf{Task}                \\ 
        \midrule
        \textbf{ShARC}~\cite{ShARC18}          & 32k           & deductive                 & government document         & conversation + boolean QA    \\ 
        \hline
        \textbf{ROPES}~\cite{ROPES19}          & 14k           & deductive                 & science textbook, Wikipedia & RC + extractive QA                \\ 
        \hline
        \textbf{ARC}~\cite{ARC18}            & 2k            & abductive                 & news comment                & 2-choice                     \\ 
    \bottomrule
    \end{tabular}
    \caption{\label{tab:other_data}Some other NLP benchmarks requiring natural language reasoning.}
\end{table}

\section{Discussion}\label{sec:discuss}
In this section, we propose some open questions, introduce some limitations, and suggest some future directions for reasoning. Among these, we also discuss the limitations of ChatGPT and GPT4.

\subsection{Open questions} 
We propose some open questions towards the reasoning capabilities of LLMs. There are many mysteries in their emergent reasoning capabilities.
\begin{itemize}
    \item \textbf{Why are CoT prompting effective?}. Why can just produce reasoning paths, which can even be wrong, before the final answer bring such significant improvement? And why CoT prompting is only effective for LLMs? What happens to LLMs when prompting with CoT but fails at medium-size PLMs?
    
    \item \textbf{Where are these reasoning capabilities of LLMs from?}. Why can LLMs emerge reasoning capabilities just as the model size increases? Where does the magic ``Let's think step by step'' come from? How can they learn these capabilities? While the mechanism of another LLMs magic, in-context-learning, has been studied~\cite{ICL-linear22,Meta-Optimizers22,ICL-baye22}, it remains more mysterious about reasoning capabilities~\cite{why-cot22}. 
    
    \item \textbf{Do even larger models reason better?}. If LLMs can emerge reasoning capabilities that can be elicited by prompts, whether they can learn competitive reasoning capabilities just as the model size increases? Or, whether it is still beneficial to build more datasets and design reasoning algorithms?
\end{itemize}

\subsection{Limitations} 
We introduce both limitations of the current research and intrinsic in PLMs.

Firstly, there are gaps in defeasible reasoning and reasoning path evaluation.
\begin{itemize}
    
    \item \textbf{Research gap on defeasible reasoning}. While defeasible reasoning is widely used in our daily life, this topic is still under-explored in NLP. ~\cite{chatgpt_eva23} found that it is more challenging for ChatGPT to perform abductive reasoning and inductive reasoning than deduction, among which induction is the much more difficult one. 
    
    \item \textbf{Lack of effective ways to evaluate reasoning paths}. It is still challenging to automatically evaluate generated reasoning paths without ground truth. Evaluating reasoning paths might become increasingly important to build explainable and reliable AI systems, especially when more people contact and use ChatGPT-like products nowadays. 
\end{itemize}

Secondly, there are also limitations intrinsic to PLMs.
\begin{itemize}
    \item \textbf{Soft deduction can produce invalid conclusions}. Transformers can only predict conclusions with probability, irrespective of whether the conclusion of deductive reasoning is necessarily true in nature, which might prevent it from precise reasoning. This characteristic can result in a sub-optimal solution to deductive problems (including arithmetic reasoning and symbolic reasoning). For example, while ChatGPT is impressive on reasoning tasks, it still fails to achieve perfect performance on the simplest one-step deductive inference task~\cite{chatgpt_eva23}.
    
    \item \textbf{Biases on content}. PLMs make their prediction based on context. While LLMs have made huge progress in reasoning, ~\cite{human-like22} found that LLMs are biased by content like humans when performing deduction. For example, they perform worse in abstract or counterfactual situations than the realistic ones. Such biases will hinder them from actual reasoning and lead to wrong answers, degrading downstream performance. More severely, it might cause harmful societal influences due to some social biases such as gender, which also exist in GPT4~\cite{sparks23}.
    
\end{itemize}

\subsection{Future} 
We suggest some potential research directions at both the holistic and technical levels in the future. 

At the holistic level, firstly, reasoning should be generalized to more complex settings (longer steps and defeasible reasoning) and more diverse knowledge mediums (languages and modalities). Secondly, it should put more attention on interpretability and faithfulness. We introduce these directions as the following.
\begin{itemize}
    \item \textbf{Generalization to longer steps}. The multi-step performance degrades as PLMs encounter samples that require more reasoning steps than those in training data or few-shot exemplars. Although there is research on decoupled one-step inference, which can alleviate the challenge of the OOD problem, it still struggles with planning. How to better generalize to longer steps is an important problem for complex reasoning tasks, which are also challenging to ChatGPT~\cite{chatgpt_eva23}.
    
    \item \textbf{More researches on defeasible reasoning}. PLMs are currently the most potential path to defeasible reasoning due to advantages we have introduced in Sec~\ref{sec:why}. According to philosophy, non-deductive reasoning is much more common than deductive reasoning in our daily lives and practical scenes. It is worth more effort to explore PLMs on defeasible reasoning since there lack of effective methods to deal with defeasible reasoning, while deductive reasoning can be solved by developed symbolic engines\footnote{Arithmetic reasoning and symbolic reasoning, which are popular recently, are also deductive and can be solved by calculator or code.}, e.g. Prolog coding for first-order logic. Moreover, it might benefit scientific research a lot if AI can induce general rules from specific facts.
    
    
    \item \textbf{Reasoning over non-English languages}. In addition to reason over English statements, it is also important to perform reasoning with other languages, which is much more challenging due to more severe data sparsity problems.
    
    \item \textbf{Reasoning with multi-modality}. Other types of modalities can also contribute to reasoning, such as tables~\cite{TabFact20} and images~\cite{NLVR19,DVD21,GD-VCR21,PMR22}. Recently, GPT4 can process images, which might push forward visual reasoning.
    
    \item \textbf{Interpretability and faithful reasoning}. Transparent and reliable reasoning paths become increasingly important when it generalizes to longer steps and defeasible reasoning. Firstly, when there are many steps, it takes more time and effort for people to check the quality of reasoning. Therefore, unfaithful reasoning might introduce difficulty in people's judgement and decision-making. Secondly, when it comes to defeasible reasoning, exposing interpretable reasoning paths is much more important and sometimes necessary for people to be convinced. In this case, different people with different background knowledge can derive different and even opposed conclusions, thus it is crucial to illustrate the evidence collected to reason.
\end{itemize}

At the technical level, we suggest several directions to improve reasoning capabilities and performance of multi-step reasoning and defeasible reasoning as follows.

\begin{itemize}
    \item \textbf{More prompts to elicit reasoning capabilities from LLMs}. Few-shot CoT and zero-shot CoT prompting are inspiring, and CoT annotations have been used to improve LLMs' reasoning capabilities~\cite{Flan22}. It is both interesting and important to find whether there are other prompts that can activate LLMs to perform reasoning or are beneficial to improving reasoning capabilities, especially on complex reasoning.
    
    \item \textbf{Self-improvement of LLMs}. Data annotations for reasoning paths, especially for long-step and defeasible reasoning, are difficult to obtain. Interestingly, in our case studies, we found that ChatGPT can provide more comprehensive answers than the ground annotations in some existing datasets such as EntailmentBank~\cite{EntailmentBank21} and WikiWhy~\cite{WikiWhy22}. Recent research have demonstrated that LLMs can learn from their self-generated reasoning paths to improve reasoning capabilities~\cite{STaR22,self-improve22}, which is the potential to alleviate the data challenge.
    
    \item \textbf{More exploration on backward reasoning}. Backward reasoning can benefit both medium-size PLMs and LLMs, while CoT prompting only benefits LLMs. Moreover, it is more efficient than forward reasoning with a smaller search space, which can bring more benefits as the depth of reasoning increases. To solve more complex reasoning problems, it is worth conducting more exploration on this direction.
    
    \item \textbf{More researches on planning}. Planning is important to perform longer-step reasoning since the search space will become bigger as the depth increases. 
    
    \item \textbf{Exploration on self-correction}. Since the conclusion of defeasible reasoning can be retracted by newly added evidence, it might be important for PLMs to self-correct their conclusions as the reasoning proceeds.
\end{itemize}

\begin{acks}
We appreciate the assistance from Ridong HAN during the investigation and the suggestion from Zhihong CHEN on the figure demonstration in this survey.
\end{acks}

\bibliographystyle{abbrv}
\bibliography{ref}
\end{document}